# Synchronization Clustering based on a Linearized Version of Vicsek model

Xinquan Chen[1, 2]

[1]Web Sciences Center, University of Electronic Science & Technologhy of China, China

[2]School of Computer Science & Engineering, Chongqing Three Gorges University, China

chenxqscut@126.com

**Abstract**: This paper presents a kind of effective synchronization clustering method based on a linearized version of Vicsek model. This method can be represented by an Effective Synchronization Clustering algorithm (ESynC), an Improved version of ESynC algorithm (IESynC), a Shrinking Synchronization Clustering algorithm based on another linear Vicsek model (SSynC), and an effective Multi-level Synchronization Clustering algorithm (MSynC). After some analysis and comparisions, we find that ESynC algorithm based on the Linearized version of the Vicsek model has better synchronization effect than SynC algorithm based on an extensive Kuramoto model and a similar synchronization clustering algorithm based on the original Vicsek model. By simulated experiments of some artificial data sets, we observe that ESynC algorithm, IESynC algorithm, and SSynC algorithm can get better synchronization effect although it needs less iterative times and less time than SynC algorithm. In some simulations, we also observe that IESynC algorithm and SSynC algorithm can get some improvements in time cost than ESynC algorithm. At last, it gives some research expectations to popularize this algorithm.

## 1. Introduction

Clustering is an unsupervised learning method that tries to find some obvious distribution structures and patterns in unlabeled data sets by maximizing the similarity of the objects in a common cluster and minimizing the similarity of the objects in different clusters.

The traditional clustering algorithms are usually categorized into partitioning methods, hierarchical methods, density-based methods, grid-based methods, model-based methods, and graph-based methods. Recent clustering methods have quantum clustering algorithms, spectral clustering algorithms, and synchronization clustering algorithms.

This paper researchs another synchronization clustering method based on a linearized version of Vicsek model comparing to SynC algorithm, which is a famious synchronization clustering algorithm presented in [1]. The major contributions of this paper can be summarized as follows:

(1) It presents an Effective Synchronization Clustering algorithm (ESynC), which is an improved version of SynC algorithm, by using a linearized version of the Vicsek model. It also presents an Improved version of ESynC algorithm (IESynC), a Shrinking Synchronization Clustering algorithm based on another linear Vicsek model (SSynC), and an effective Multi-level Synchronization Clustering algorithm (MSynC).

(2) It analyzes the reason why the linearized version of the Vicsek model is more effective than the extensive Kuramoto model for exploreing clusters and outliers in dynamic clustering process.

(3) It validates the improved effect of MSynC algorithm by the simulated experiments of several different kinds of data sets.

The remainder of this paper is organized as follows. Section 2 lists some related papers. Section 3 gives some basic concepts. Section 4 introduces ESynC algorithm. Section 5 introduces SSynC algorithm. Section 6 introduces MSynC algorithm. Section 7 validates our algorithms by some simulated experiments. Conclusions and future works are presented in Section 8.

## 2. Related Work

This paper is inspired by several papers [1, 2, 3, 4]. In 2010, Böhm et al. [1] presented a novel clustering approach, SynC algorithm, inspired by the synchronization principle. SynC algorithm can find the intrinsic structure of the data set without any distribution assumptions and handle outliers by dynamic synchronization. In order to implement automatic clustering, those natural clusters can be discovered by using the Minimum Description Length principle (MDL) [5].

In 1995, Vicsek et al. [2] presented a basic model of multi-agent systems that contains noise effects. This basic model can also be regarded as a special version of Reynolds model [6]. Simulation results demonstrate that some systems using Vicsek model [2] or one-dimensional models presented by Czirok et al. [7] can be synchronized when they has large population density and small noise. Naturally, we expect that this kind of model can be used to explore clusters and noises of some data sets by local synchronization.

In 2003, Jadbabaie et al. [3] analyzed a simplified Vicsek model without noise effects and provided a theoretical explanation for the nearest neighbor rule that can cause all agents to eventually move in the same direction.

In 2008, Liu et al. [8] provided the synchronization property of the Vicsek model after given initial conditions and the model parameters and revealed some fundamental differences between the Vicsek model and its linearized version. In 2009, Wang et al. [4] researched Vicsek model under noise disturbances and prosented some theoretical results.

After 2010, J. Shao et al. published several synchronization clustering papers from several views [9, 10, 11, 12]. In order to detect the outliers from a real complex data set more naturally, a novel outlier detection algorithm was presented from a new perspective, "Out of Synchronization" [9]. In order to find subspace clusters of some high-dimensional sparse data sets, a novel effective and effcient subspace clustering algorithm, ORSC [10], was proposed. In order to explore meaningful levels of the hierarchical cluster structure, a novel dynamic hierarchical clustering algorithm, hSync [11], was presented based on synchronization and the MDL principle. In order to find the intrinsic patterns of a complex graph, a novel and robust graph clustering algorithm, RSGC [12], was proposed by regarding the graph clustering as a dynamic process towards synchronization. In 2013, JianBin Huang et al. [13] also presented a synchronization-based hierarchical clustering method basing on the work of [1].

Recent years, some physicists also research the explosive synchronization in some complexity networks to uncover the underlying mechanisms of the synchronization state [14, 15, 16]. In these papers, the synchronization rules in some networks are explored.

## 3. Some Basic Concepts

Suppose there is a data set $S = \{X_1, X_2, \ldots, X_n\}$ in a $d$-dimensional Euclidean space. Naturally, we use Euclidean metric as our dissimilarity measure, $dis(\cdot, \cdot)$. In order to describe our algorithm clearly, some concepts are presented first.

**Definition 1**. The $\delta$ near neighbor point set $\delta(P)$ of point $P$ is defined as

$$\delta(P) = \{X \mid dis(X, P) \leq \delta, X \in S, X \neq P\}, \tag{1}$$

where $dis(X, P)$ is the dissimilarity measure between point $X$ and point $P$ in the data set $S$. Parameter $\delta$ is a predefined threshold.

**Definition 2** [1]**.** The extensive Kuramoto model for clustering is defined as

Point $X = (x_1, x_2, \ldots, x_d)$ is a vector in $d$-dimensional Euclidean space. If each point $X$ is regarded as a phase oscillator according to Eq.(1) of [1], with an interaction in the $\delta$ near neighbor point set $\delta(X)$, then the dynamics of the $k$ dimension $x_k$ ($k = 1, 2, \ldots, d$) of point $X$ over time is described by:

$$x_k(t+1) = x_k(t) + \frac{1}{|\delta(X(t))|} \sum_{Y \in \delta(X(t))} \sin(y_k(t) - x_k(t)), \tag{2}$$

where $X(t = 0) = (x_1(0), x_2(0), \ldots, x_d(0))$ represents the original phase of point $X$, and $x_k(t+1)$ describes the renewal phase value in the $k$-th dimension of point $X$ at the $t$ step evolution.

**Definition 3** [17]. The $t$-step $\delta$ near neighbor undirected graph $G_\delta(t)$ of the data set $S = \{X_1, X_2, \ldots, X_n\}$ is defined as

$$G_\delta(t) = (V(t), E(t)), \tag{3}$$

where $V(t = 0) = S = \{X_1, X_2, \ldots, X_n\}$ is the original vertex set, $E(t = 0) = \{(X_i, X_j) \mid X_j \in \delta(X_i), X_i\ (i = 1, 2, \ldots, n) \in S\}$ is the original edge set. $V(t) = \{X_1(t), X_2(t), \ldots, X_n(t)\}$ is the $t$-step vertex set of $S$, $E(t) = \{(X_i(t), X_j(t)) \mid X_j(t) \in \delta(X_i(t)), X_i(t)\ (i = 1, 2, \ldots, n) \in V(t)\}$ is the $t$-step edge set, and the weight-computing equation of edge $(X_i, X_j)$ is $weight(X_i, X_j) = dis(X_i, X_j)$.

**Definition 4** [17]. The $t$-step average length of edges, $AveLen(t)$, in a $t$-step $\delta$ near neighbor undirected graph $G_\delta(t)$ is defined as

$$AveLen(t) = \frac{1}{|E(t)|} \sum_{e \in E(t)} |e|, \tag{4}$$

where $E(t)$ is the $t$-step edge set, and $|e|$ is the length (or weight) of edge $e$. The average length of edges in $G_\delta(t)$ decreases to its limit 0, that is $AveLen(t) \to 0$, as more $\delta$ near neighbor points synchronize together with time evolution.

**Definition 5** [1]. The cluster order parameter $r_c$ characterizing the degree of local synchronization is defined as:

$$r_c = \frac{1}{n} \sum_{i=1}^{n} \sum_{Y \in \delta(X)} e^{-dis(X,Y)}. \tag{5}$$

**Definition 6.** The Vicsek model [2, 3] for clustering is defined as

Point $X = (x_1, x_2, \ldots, x_d)$ is a vector in $d$-dimensional Euclidean space. If each point $X$ is regarded as an agent according to the Vicsek model [2, 3], with an interaction in the $\delta$ near neighbor point set $\delta(X)$, then the dynamics of point $X$ over time according to [2, 3] is described by:

$$X(t+1) = X(t) + \frac{X(t) + \sum_{Y \in \delta(X(t))} Y}{\left\| X(t) + \sum_{Y \in \delta(X(t))} Y \right\|} \cdot v(t) \cdot \Delta t, \tag{6}$$

where $X(t = 0) = (x_1(0), x_2(0), \ldots, x_d(0))$ represents the original location of point $X$, $X(t+1)$ describes the renewal location of point $X$ at the $t$ step evolution, $v(t)$ is the move velocity at the $t$ step evolution, and $v(t) \cdot \Delta t$ is the move length at the $t$ step evolution.

A special case of this original version of the Vicsek model is that if the $\delta$ near neighbor point of one point is null, this point still will move along its vector direction.

In some multi-agent systems based on the Vicsek model, $v(t)$ is a constant. If $v(t)$ is always a constant, maybe Eq. (6) can not be used for clustering. In a simulation using the data set of Fig.1, we find that the original version of the Vicsek model based on Eq. (6) can not work well for clustering when $v(t)$ is a constant. So we present another effective version of the Vicsek model for clustering.

**Definition 7.** A linearized version of Vicsek model for clustering is defined as

Point $X = (x_1, x_2, \ldots, x_d)$ is a vector in $d$-dimensional Euclidean space. If each point $X$ is regarded as a phase oscillator according to a linearized version of Vicsek model [2], with an interaction in the $\delta$ near neighbor point set $\delta(X)$, then the dynamics of point $X$ over time according to [3] and [4] is described by:

$$X(t+1) = \frac{1}{\left(1 + |\delta(X(t))|\right)} \left( X(t) + \sum_{Y \in \delta(X(t))} Y \right), \tag{7}$$

where $X(t = 0) = (x_1(0), x_2(0), \ldots, x_d(0))$ represents the original phase of point $X$, and $X(t+1)$ describes the renewal phase value of point $X$ at the $t$ step evolution.

Eq.(7) can also be rewritten by:

$$\begin{aligned} X(t+1) &= X(t) + \sum_{Y \in \delta(X(t))} \left(Y - X(t+1)\right) \\ &= X(t) + \frac{1}{\left(1 + |\delta(X(t))|\right)} \left( \sum_{Y \in \delta(X(t))} \left(Y - X(t)\right) \right), \end{aligned} \tag{8}$$

Eq.(8) has some similarity with Eq.(2) in form, but they have essential difference. We can ses that the renewal mode of Eq.(2) is nonlinear and the renewal mode of Eq.(7) and Eq.(8) is linear.

A special case of this linear Vicsek model is that two points satisfied that the $\delta$

near neighbor point of one point only contains another point. After one time synchronization using Eq.(7), the two points will be superposed to their middle location. So we think this model is more consistent of the intuition for clustering. Another special case is that one point satisfied that the $\delta$ near neighbor point of the point is null. After one time synchronization using Eq.(7), the point is still an isolate.

**Definition 8**. Accordint to [18], the Grid Cell is defined as follows:

Grid cells of the data space can be obtained after partitioned the multidimensional ordered-attribute space using a kind of multidimensional grid partitioning method.

The data structure of grid cell $g$ can be defined as:

$$DS(g) = (\text{Grid_Label, Grid_ Coordinate, Grid_Location, Grid_Range, Point_Number, Points_Set}). \quad (9)$$

In Eq.(9),

Grid_Label is the key label of the grid cell.

Grid_Coordinate is the coordinate of the grid cell. It is a $d$-dimensional integer vector expressed by $I = (i_1, i_2, \ldots, i_d)$.

Grid_Location is the center location of the grid cell. It is a $d$-dimensional vector expressed by $P = (p_1, p_2, \ldots, p_d)$.

Grid_Range records the region of the grid cell. It is a $d$-dimensional interval vector expressed by:

$$R = ([p_1\text{-}r_1/2, p_1+r_1/2), \ldots , [p_d\text{-}r_d/2, p_d+r_d/2)), \quad (10)$$

where $r_i$ ($i$ =1, 2, …, $d$) is the interval length in the $i$-th dimension of the grid cell.

Point_Number records the number of points of the grid cell.

Points_Set records the labels of points of the grid cell.

In FSynC algorithm [17] and our IESynC algorithm, we use a Red-Black tree to records the labels of points of the grid cell to obtain efficient inserting and deleting operations. In this paper, Grid cells and Red-Black trees can be used to decrease the time cost of ESynC algorithm.

**Definition 9**. The Core is defined as follows:

The data structure of core $C$ can be defined as:

$$DS(C) = (\text{Core_Id, Core_Location, Parent_CoreId, Number_ContainingPoints}). \quad (11)$$

In Eq.(11),

Core_Id is the identification number in the original data set.

Core_Location is the current location of core $C$. It is a $d$-dimensional vector expressed by $C = (c_1, c_2, \ldots, c_d)$.

Parent_CoreId is the identification number of the parent of the core $C$. At the original of clustering, the Parent_CoreId of the core $C$ is itself. At the middle or final of clustering, the Parent_CoreId of the core $C$ is the Core_Id of the attributive core of the core $C$.

Number_ContainingPoints is the number of points that are represented or contained by the core $C$.

The data structure of core $C$ can be defined by a struct using C language or a class using C++ language.

**Definition 10.** A synchronization model for clustering a core set is defined as

Core $C = (c_1, c_2, \ldots, c_d)$ is a vector in a $d$-dimensional Euclidean space. If each core $C$ is regarded as a phase oscillator according to an extended linearized version of Vicsek model [2], with an interaction in the $\delta$ near neighbor point set $\delta(C)$, then the dynamics of the $k$ dimension $c_k$ ($k = 1, 2, \ldots, d$) of core $C$ over time is described by:

$c_k\,(t+1) =$

$$\frac{1}{(count(C(t)) + \sum_{Y\in\delta(C(t))} count(Y(t)))}\left(count(C(t))\cdot x_k(t) + \sum_{Y\in\delta(C(t))}\left(count(Y(t))\cdot y_k(t)\right)\right), \quad (12)$$

where $C(t = 0) = (c_1(0), c_2(0), \ldots, c_d(0))$ represents the original phase of core $C$, $c_k\,(t+1)$ describes the renewal phase value in the $k$-th dimension of core $C$ at the $t$ step evolution, and $count(C)$ represents the value of the Number_ContainingPoints of the core $C$.

In the dynamics clustering, if the Parent_CoreId of the core $C$ is itself and the value of the Number_ContainingPoints of the core $C$ is equal to 1, then Eq.(12) is equivalent with Eq.(7). Acturally, in the dynamics clustering, if the core $C$ is represented by its parent core (which means that the value of the Number_ContainingPoints of the parent core is added by $count(C)$), then Eq.(12) can also be used for saving time and space.

According to this dynamics clustering model represented by Eq.(12), a big data set can be partitioned into multi subsections that can be loaded group by group into the memory for clustering.

**Definition 11.** The data set $S = \{X_1, X_2, \ldots, X_n\}$ using the linearized version of Vicsek model described by Eq.(7) for clustering is said to achive local synchronization if the final locations of all points satisfy

$$X_i(t = T) = RC_k(T),\ i = 1, 2, \ldots, n,\ k = 1, 2, \ldots, K, \tag{13}$$

where $T$ is the times of the final synchronization, $K$ is the number of the root cores in the final synchronization process, $RC_k(T)$ is the $k$-th root core in the final synchronization process.

Usually, the final location of point $X_i$ ($i = 1, 2, \ldots, n$) may depends on parameter $\delta$ and the original locations of itself and other points in the data set $S$.

**Theorem 1**. The data set $S = \{X_1, X_2, \ldots, X_n\}$ using the linearized version of Vicsek model described by Eq.(7) for clustering will achive local synchronization, if the parameter $\delta$ satisfy

$$\delta_{\min} \leqslant \delta \leqslant \delta_{\max}, \tag{14}$$

Suppose $e_{\min}(\mathrm{MST}(S))$, which is also equal to $\min\{dis(X_i, X_j) | (X_i, X_j \in S) \wedge (X_i \neq X_j)\}$, is the weight of the minimum edge in the Minimum Span Tree (MST) of the complete graph of the data set $S$, and $e_{\max}(\mathrm{MST}(S))$ is the weight of the maximum edge in the MST of the complete graph of the data set $S$. If the data set $S$ has no isolate, then usually, $\delta_{\min} = e_{\min}(\mathrm{MST}(S))$, and $e_{\max}(\mathrm{MST}(S)) \leqslant \delta_{\max} \leqslant \max\{dis(X_i, X_j) | (X_i, X_j \in S) \wedge (X_i \neq X_j)\}$. If the data set $S$ has isolate, we should filtrate all isolates before using the heuristic method to set parameter $\delta$.

**Proof**: if $\delta < \delta_{\min}$, then for any $X_i$ ($i = 1, 2, \ldots, n$), there is $\delta(X_i) = \emptyset$. In this case, Any point in the data set $S$ can not synchronize with other points, so synchronization will not happen.

In another case, that is when $\delta > \delta_{\max}$, then for any $X_i$ ($i = 1, 2, \ldots, n$), there is $\delta(X_i(t)) = S - \{X_i(t)\}$. Accordint to Eq.(7), there is $X_i(t+1) = \mathrm{mean}(S)$. Here, $\mathrm{mean}(S)$ is the mean of all points in the data set $S$. Any point in the data set $S$ will synchronize with all other points, so global synchronization will happen. After one time synchronization, all points in the data set $S$ will synchronize to their mean location.

Apparently, if $\delta_{\min} \leqslant \delta \leqslant \delta_{\max}$, local synchronization will happen. And the final result of synchronization is affected by parameter $\delta$ and the original locations of all points in the data set $S$.

**Property 1**. The data set $S = \{X_1, X_2, \ldots, X_n\}$ using the linearized version of Vicsek model described by Eq.(7) for clustering will achive an effect local synchronization with some obvious clusters and isolates, if the parameter $\delta$ satisfy

$$\max\{longthestEdgeInMsf(cluster_k) \mid k = 1, 2, \ldots, K\} < \delta < \min\{dis(cluster_i, cluster_j) \mid i, j = 1, 2, \ldots, K\}, \quad (15)$$

where $longthestE\ InMsf(cluster_k)$ is the longest edge in the minimum spanning forest of the $k$-th cluster, $dis(cluster_i, cluster_j)$ is the minimum edge connecting the $i$-th cluster and the $j$-th cluster.

**Proof**: Suppose the data set $S = \{X_1, X_2, \ldots, X_n\}$ has $K$ obvious clusters. If parameter $\delta$ is larger than $\max\{longthestEdgeInMsf(cluster_k) \mid k = 1, 2, \ldots, K\}$, then data points in the same obvious cluster will be synchronized. If parameter $\delta$ is less than $\min\{dis(cluster_i, cluster_j) \mid i, j = 1, 2, \ldots, K\}$, then data points in different obvious clusters can not be synchronized.

**Note**: Synchronization clustering using another linearized version of Vicsek model described by Eq.(12) also has the same theorem and property.

## 4. An Effective Synchronization Clustering Algorithm Based on a Linearized Version of Linear Vicsek Model

This method has almost the same process as SynC algorithm except using a different dynamics clustering model, a linearized version of Vicsek mode represented by Eq.(7).

Although we use the Euclidean metric as our dissimilarity measure in this paper, the algorithm is by no means restricted to this metric and this kind of data space. If we can construct a proper dissimilarity measure in a hybrid-attribute space, the algorithm can also be used.

### 4.1 Comparing the extensive Kuramoto model with the Linear Vicsek model

Comparing Eq.(2) and Eq.(7), we can see that renewal mode of the extensive Kuramoto model at each step evolution is nonlinear and renewal mode of the linear Vicsek model at each step evolution is linear.

Figure 1 compares the tracks of points in clustering process based on the three models respectively. Figure 2 compares the relation between the number of clusters and parameter $\delta$ after finished clustering based on the three models respectively.

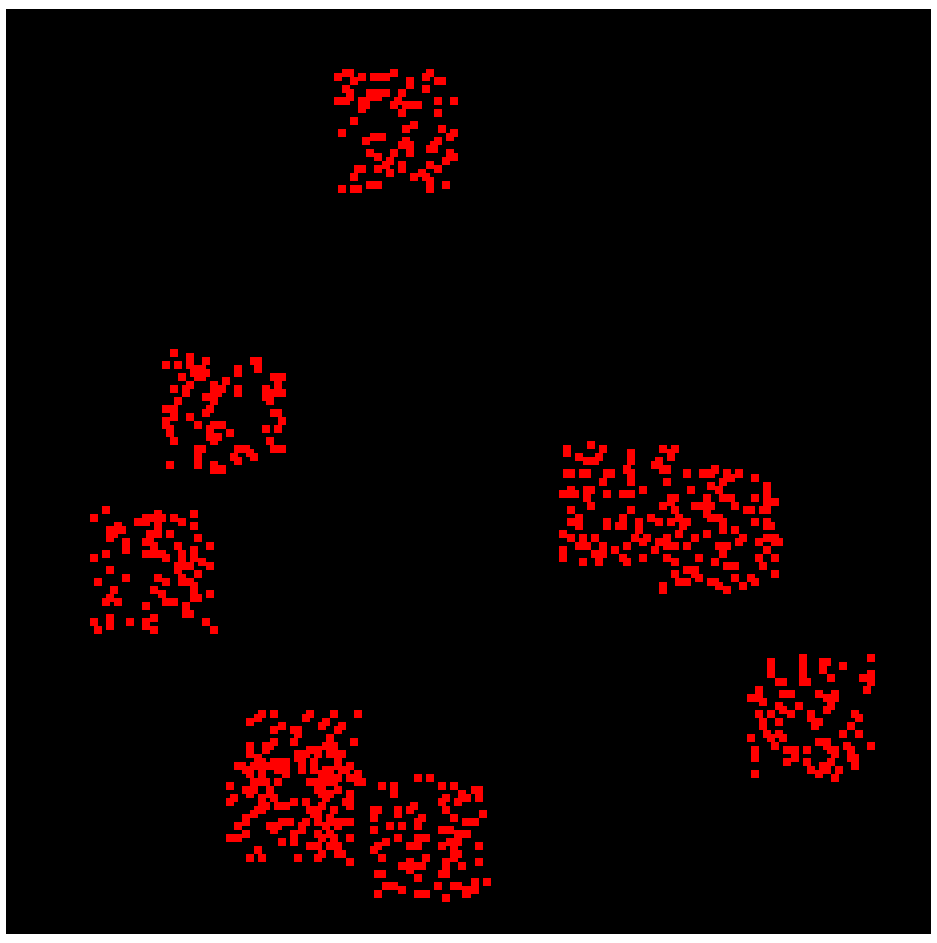

(a) $t = 0$ (The original locations of 800 data points)

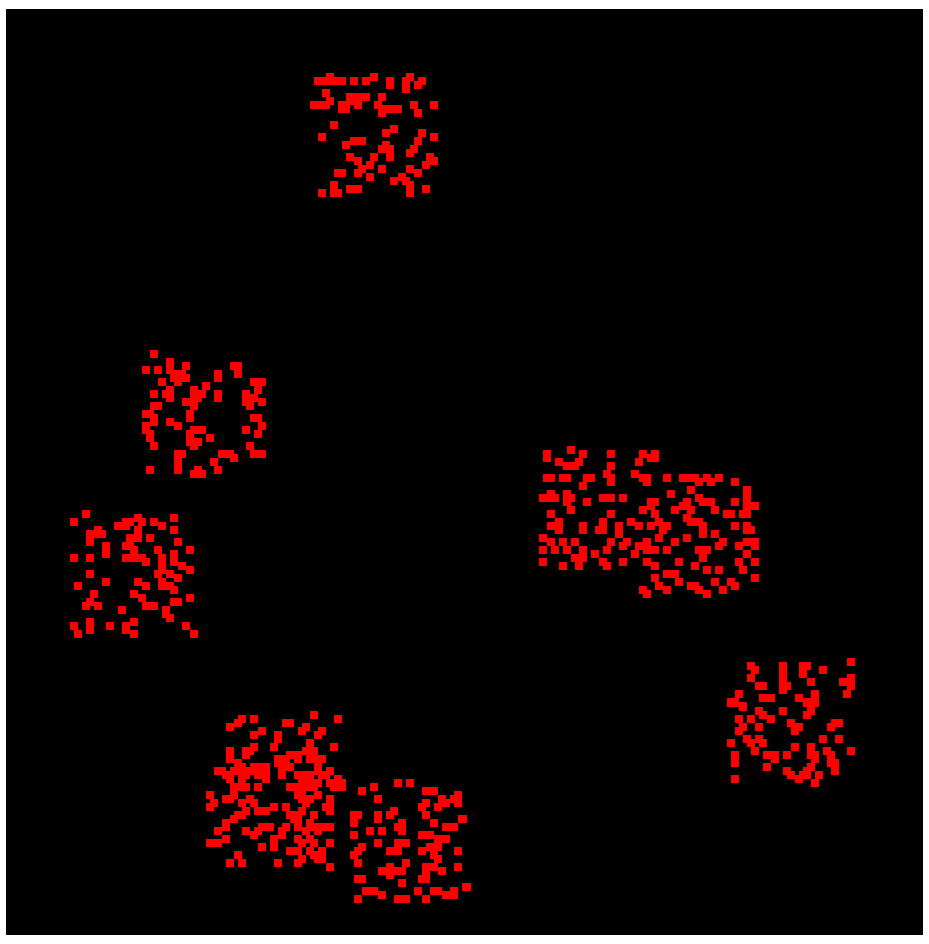

(b-1) EK Model, $t = 1$

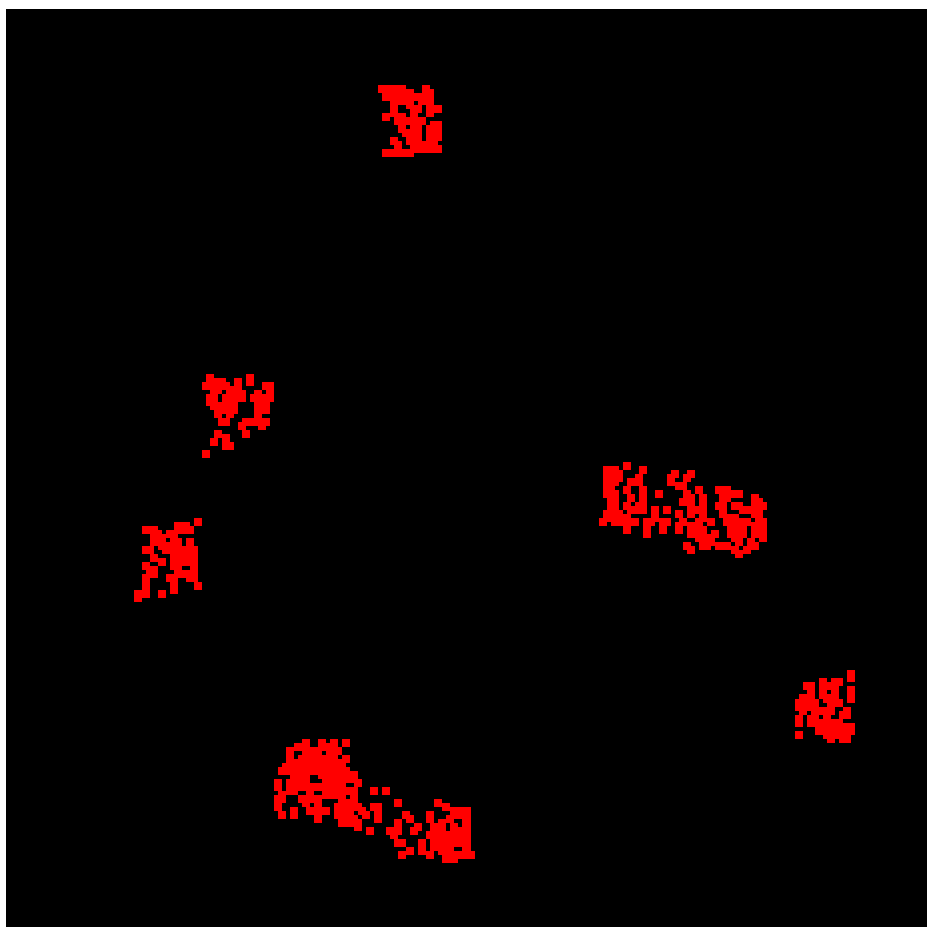

(b-2) LV Model, $t = 1$

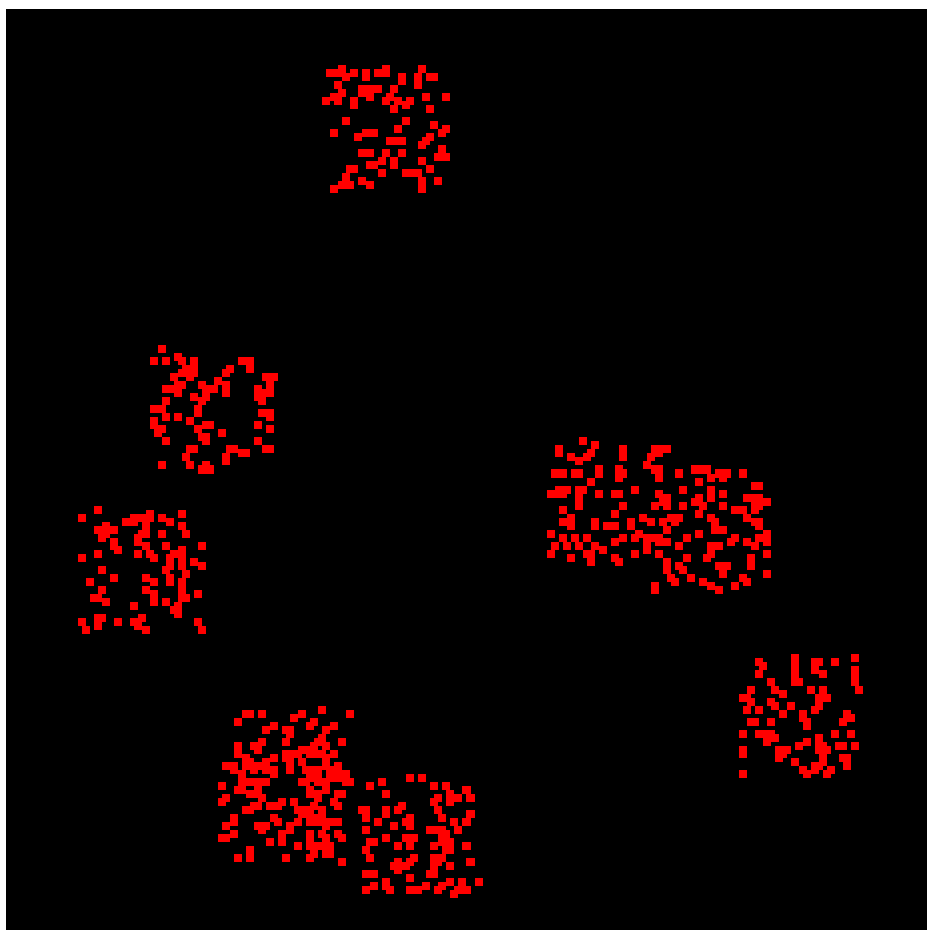

(b-3) OV Model, $t = 1$

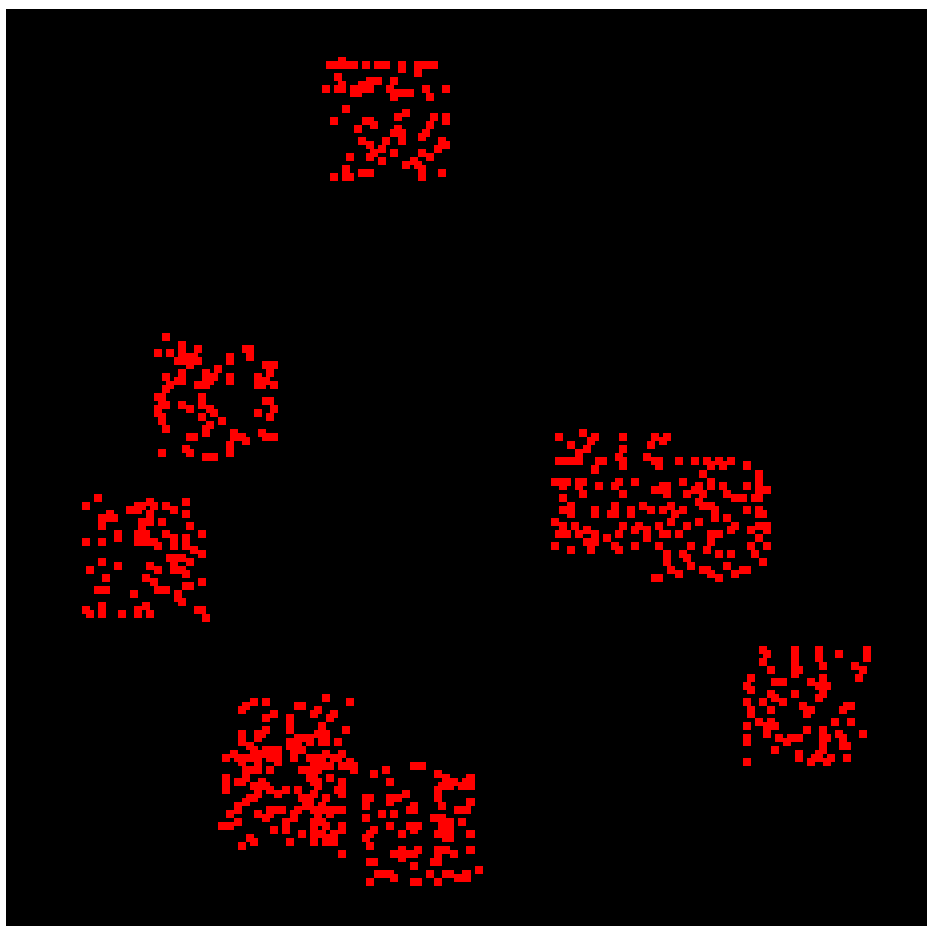

(c-1) EK Model, $t = 2$

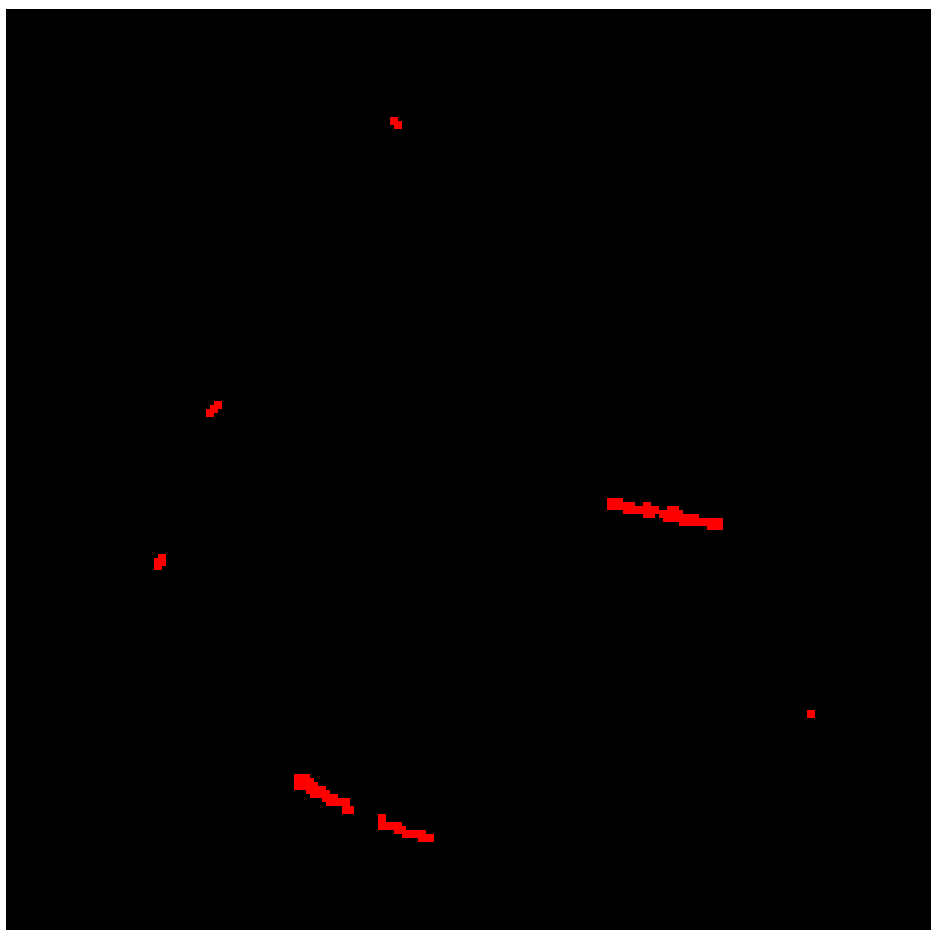

(c-2) LV Model, $t = 2$

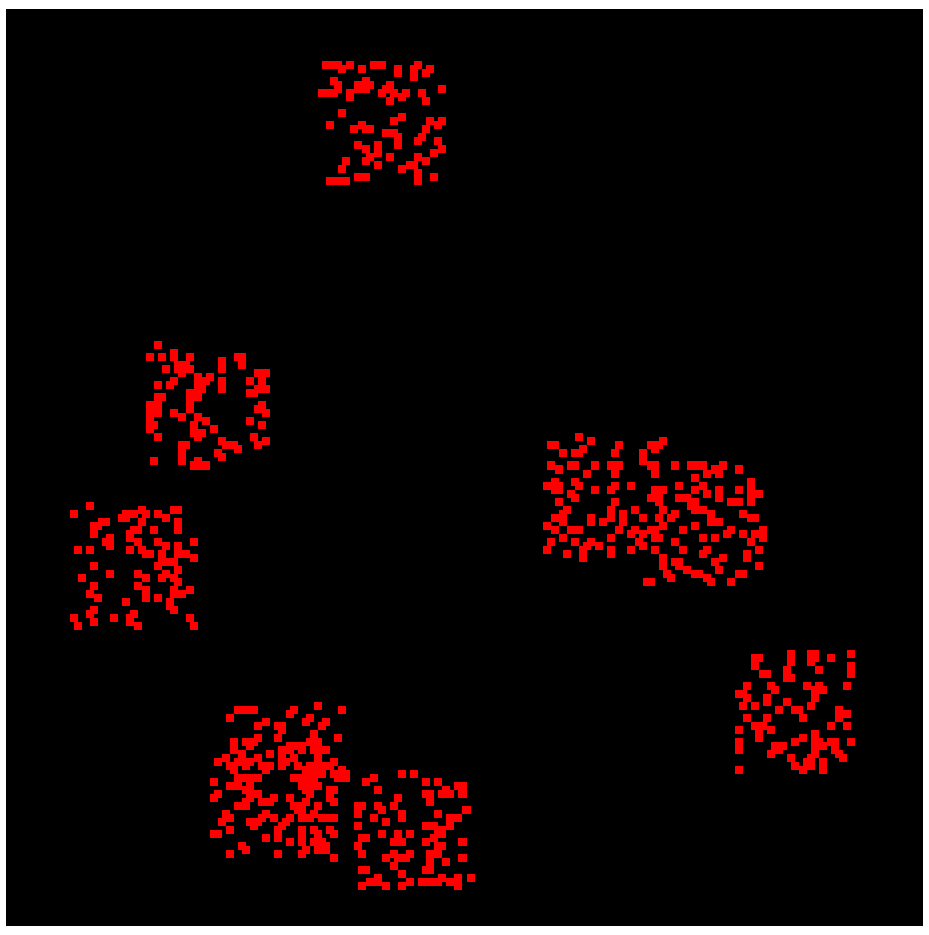

(c-3) OV Model, $t = 2$

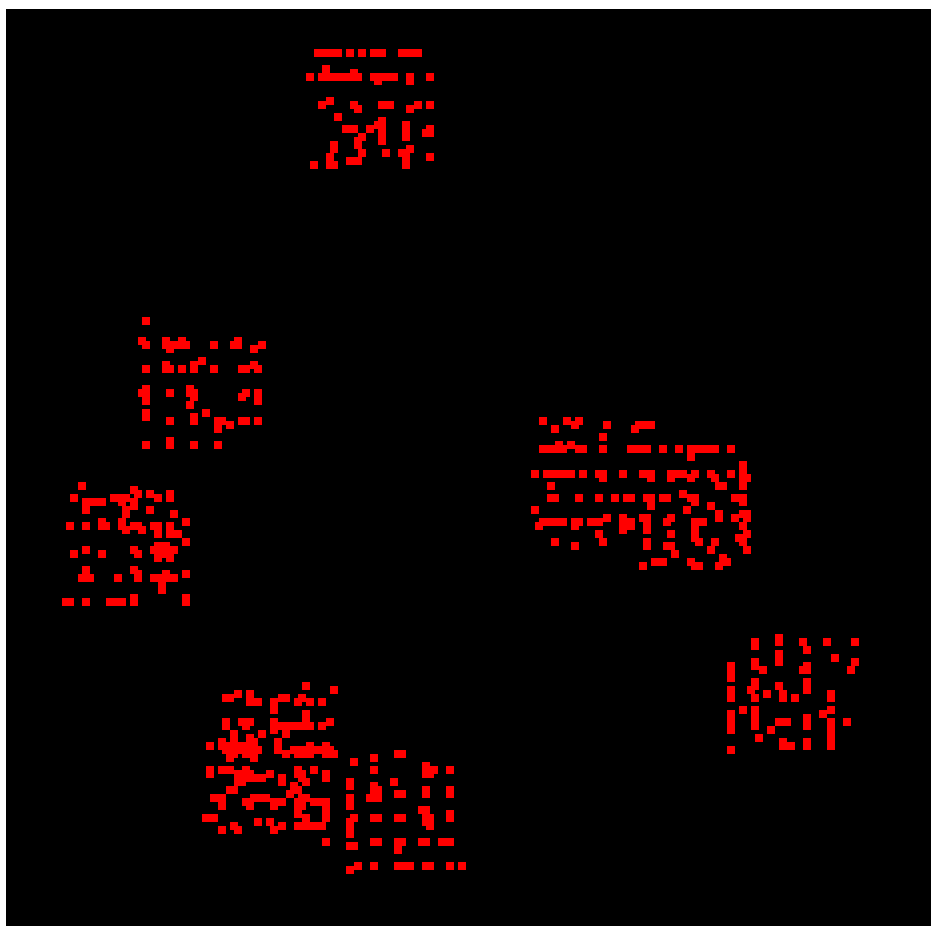

(d-1) EK Model, $t = 5$

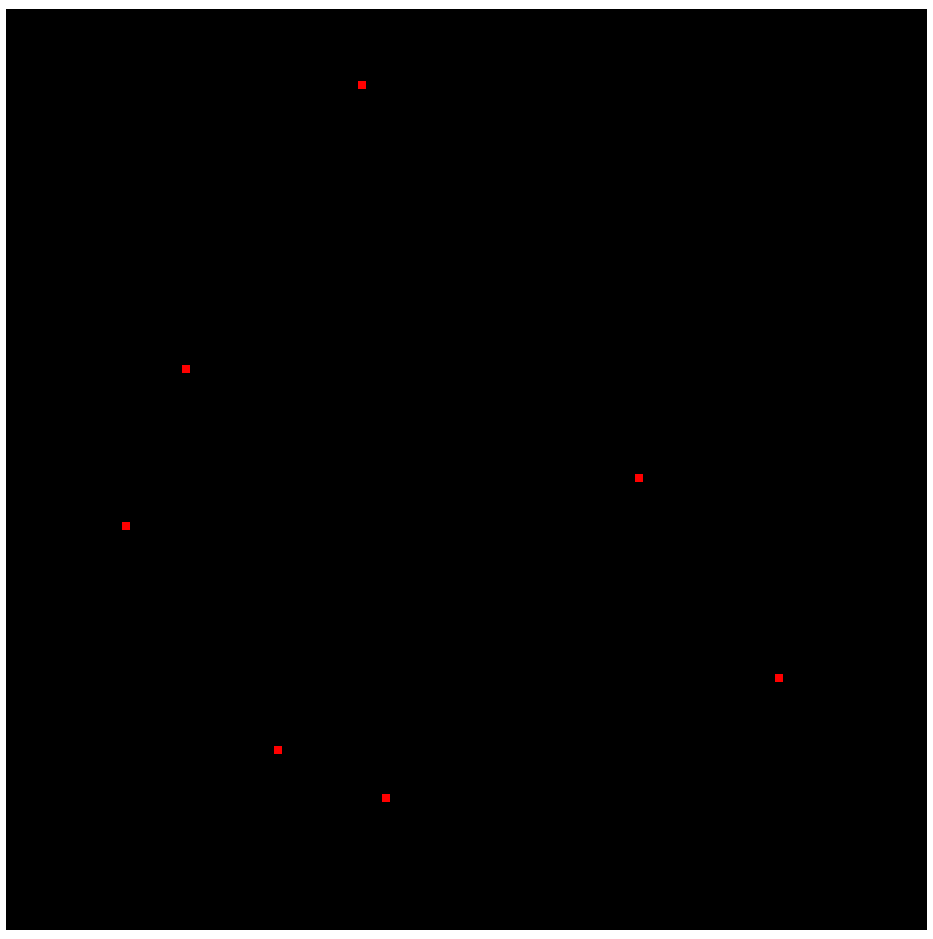

(d-2) LV Model, $t = 5$

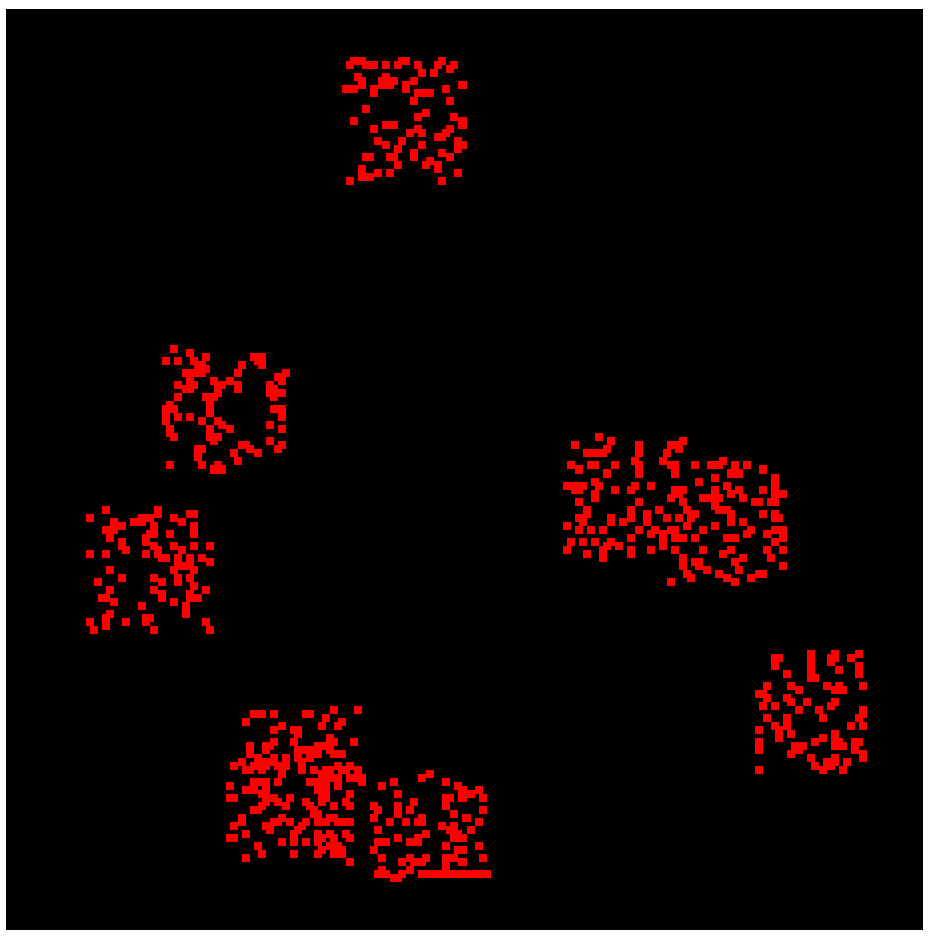

(d-3) OV Model, $t = 5$

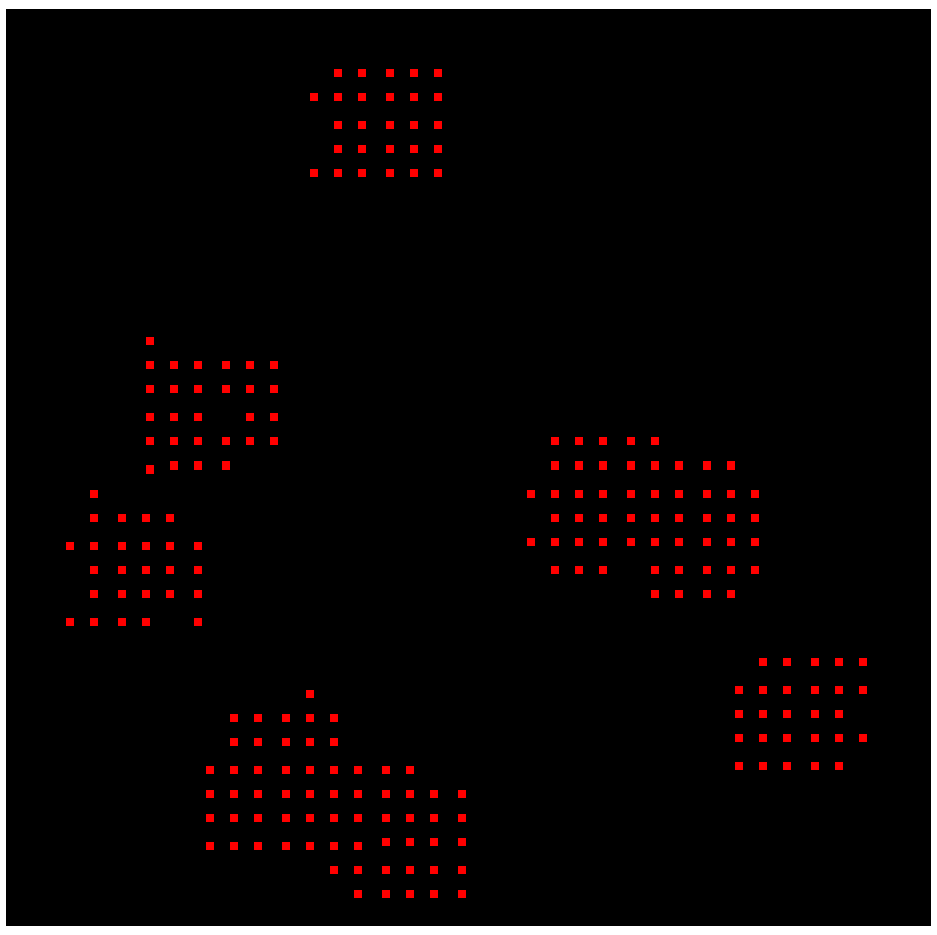

(d-1) EK Model, $t = 45$

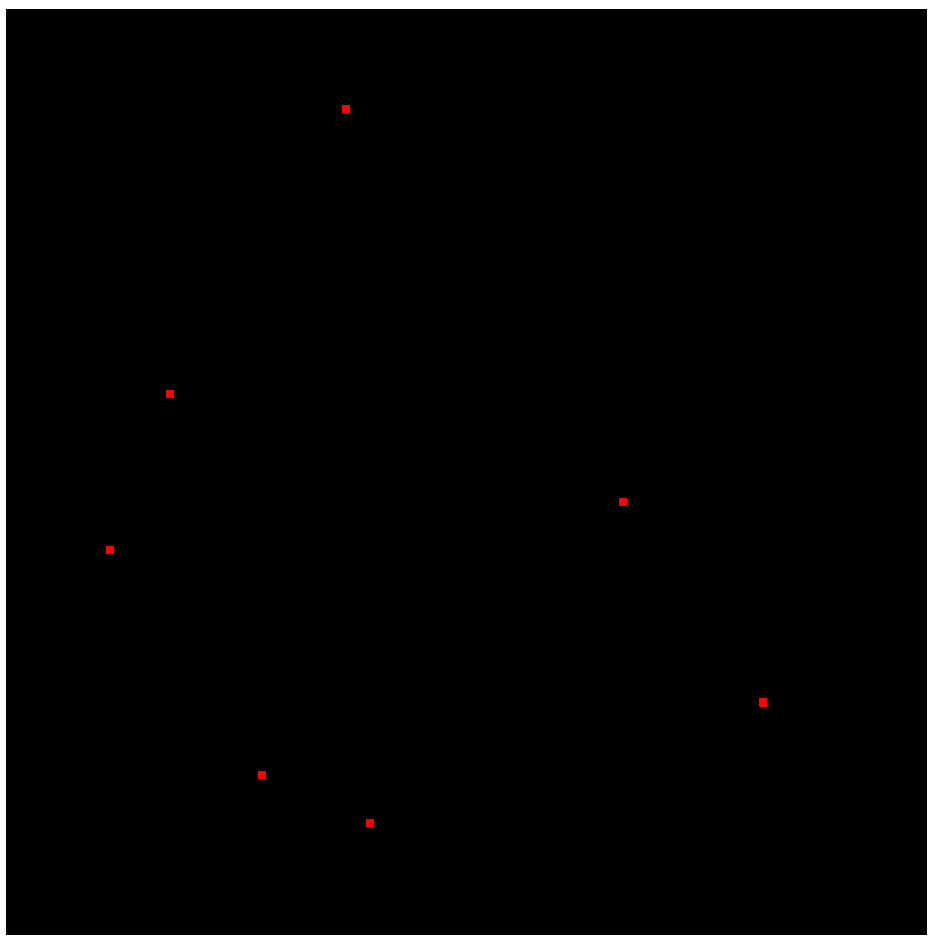

(d-2) LV Model, $t = 45$

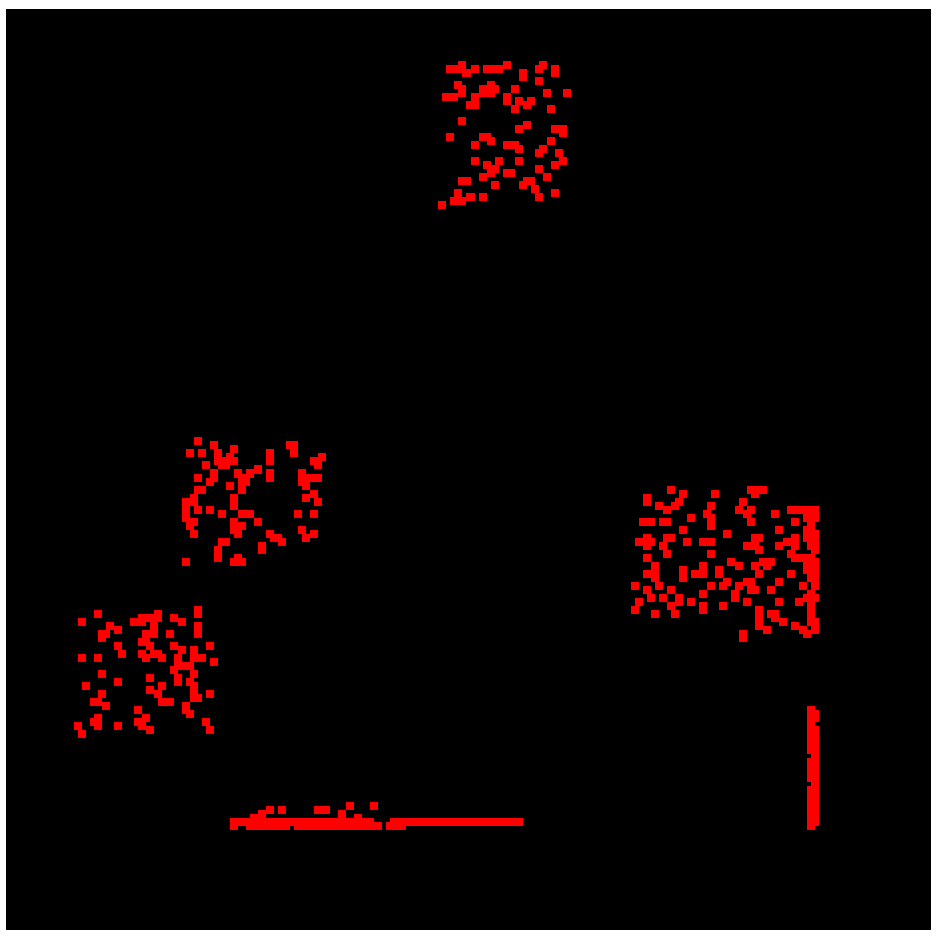

(d-3) OV Model, $t = 45$

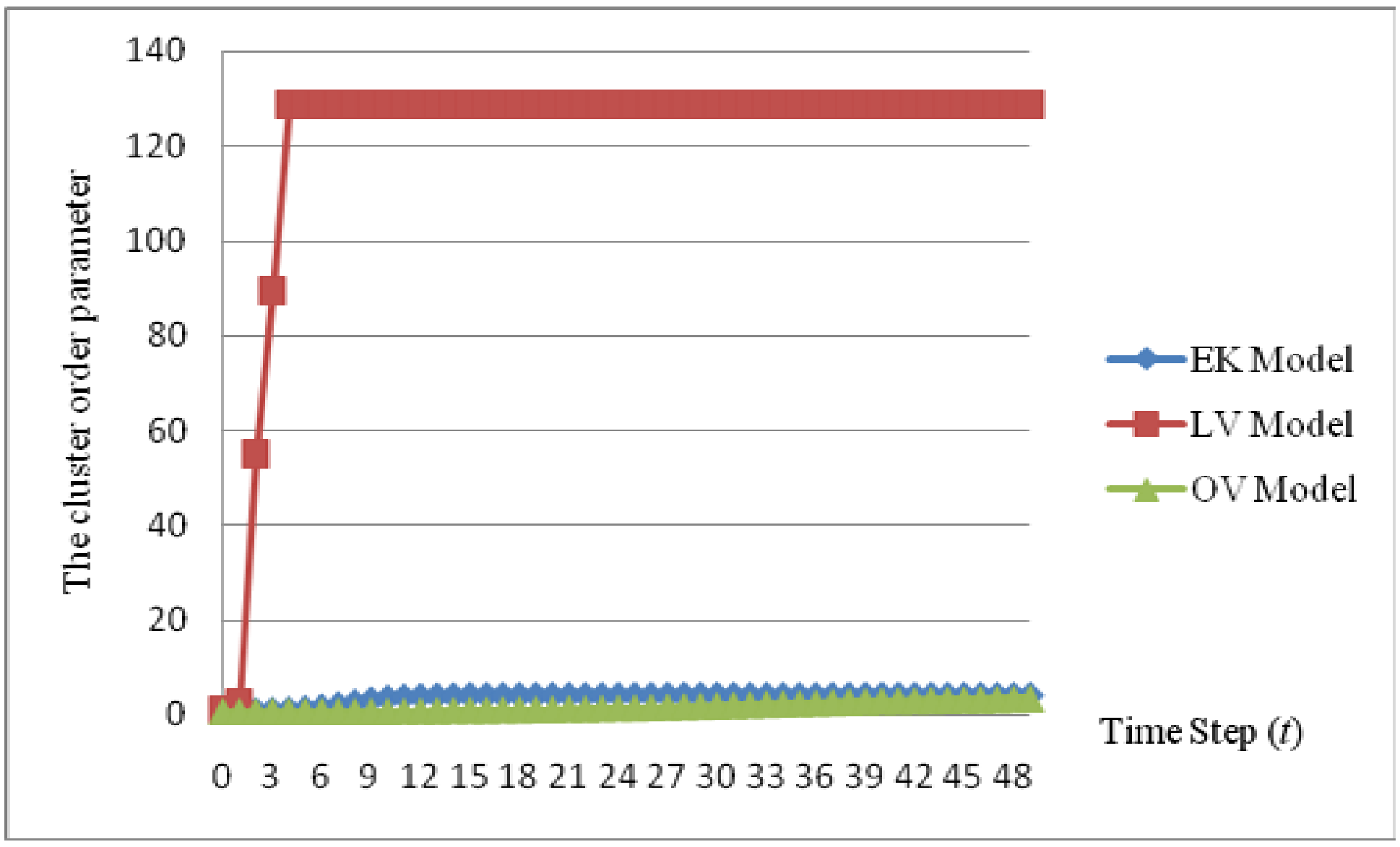

(f) The cluster order parameter with $t$-step evolution ($t$: 0 - 49)

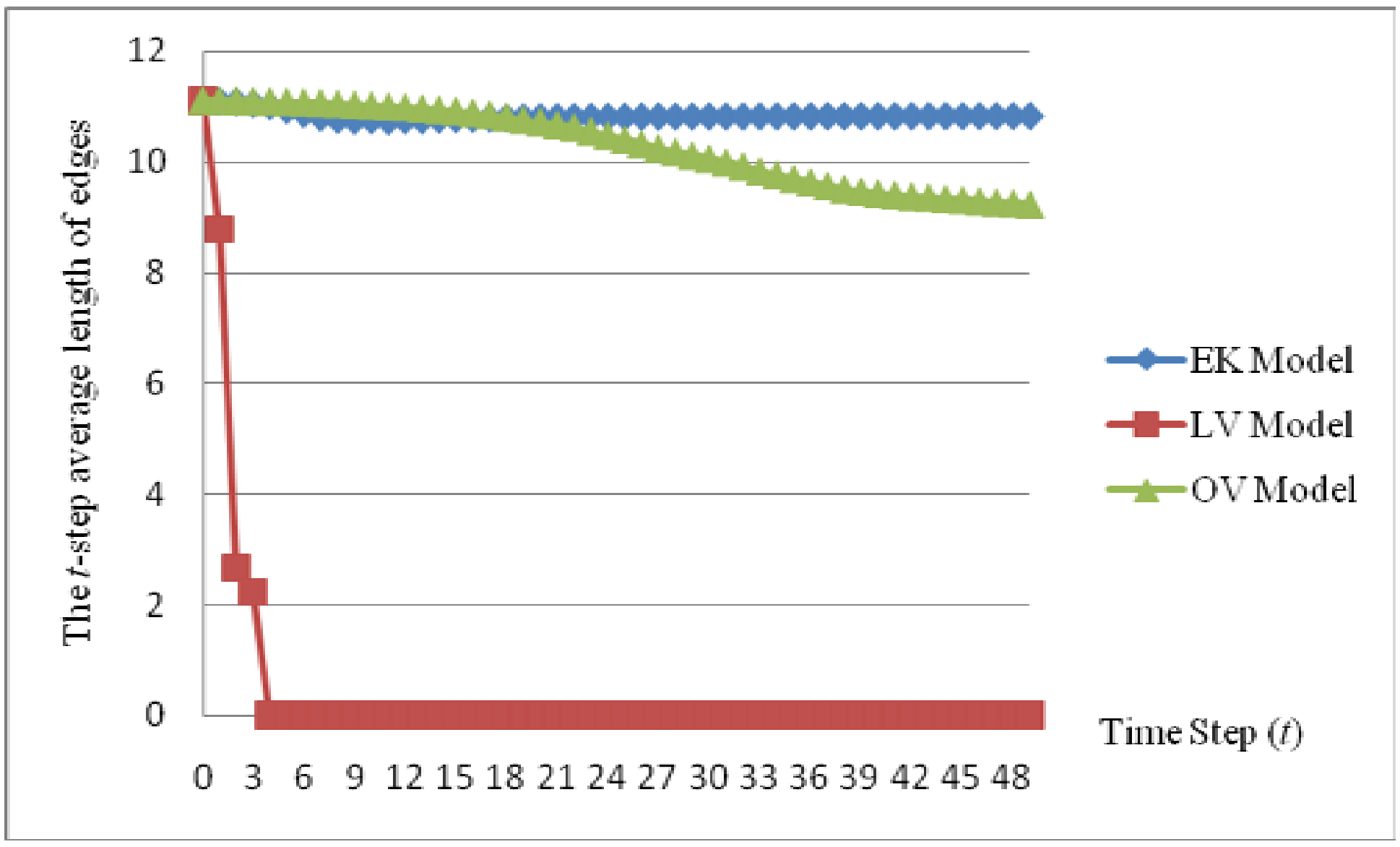

(g) The $t$-step average length of edges ($t$: 0 - 49)

Figure 1. The Process of Dynamical Clustering with Time Evolution Comparing the Extensive Kuramoto Model (EK Model) with the Linear Vicsek Model (LV Model) and the Original version of the Vicsek Model (OV Model). From (b) to (g), parameter $\delta$ is set as 18 in the three models.

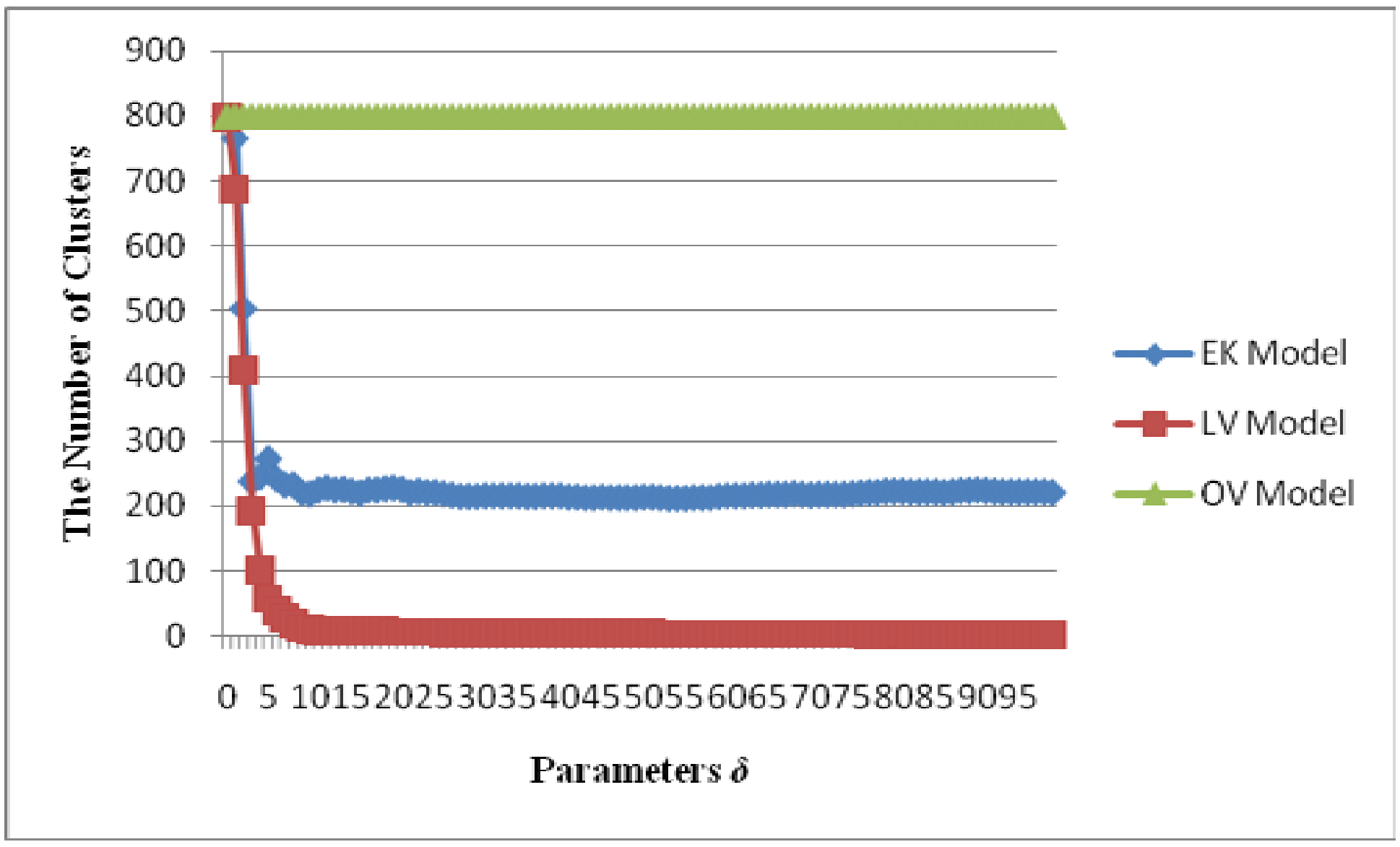

Figure 2. The relation between the number of clusters and parameter $\delta$ ($\delta$: 0 - 99). Parameter $\varepsilon$ (a very small real, if the distance of two points is less than $\varepsilon$, then they are regarded as in the same cluster) is set as 0.00001.

### 4.2 The Description of ESynC Algorithm

**Algorithm Name**: an Effective Synchronization Clustering algorithm (ESynC)

**Input**: data set $S = \{X_1, X_2, \ldots, X_n\}$, dissimilarity measure $d(\cdot, \cdot)$, and parameter $\delta$.

**Output**: The final convergent result $S(T) = \{X_1(T), X_2(T), \ldots, X_n(T)\}$ of the original data set $S$.

**Procedure**:

Step1. Initialization:

```
IterateStep t is set as zero firstly, that is: t ← 0;
for (i = 1; i ≤ n; i++)
    X_i(t) ← X_i;
```

Step2. Execute the iterative synchronization process of the dynamical clustering:

```
while ((the dynamical clustering does not satisfy its convergent condition) and (t < 50))
{
    for (i = 1; i ≤ n; i++)
    {
        Construct the δ near neighbor point set δ(X_i(t)) for each point X_i(t) (i = 1, 2, …, n) according to Definition 1;
        Compute the renewal value, X_i(t+1), of X_i(t) using Eq.(7);
    }
    Compute the t-step average length of edges of all points, AveLen(t), using Eq.(4);
    /* We can also compute the cluster order parameter r_c according to Eq.(5) of Definition 5 instead of computing AveLen(t). */
    IterateStep t is increased by 1, that is: t++;
    if (AveLen(t) → 0)    /* AveLen(t) → 0 is equivalent with r_c → 1. */
        We think the dynamical clustering reaches its convergent result, and then exit from the while repetition;
}
```

Step3. Finally we get a convergent result $S(T) = \{X_1(T), X_2(T), \ldots, X_n(T)\}$, where $T$ is the times of the while repetition in Step2. The final convergent set $S(T)$ reflects the natural clusters or isolates of the data set $S$.

### 4.2 Time Complexity Analysis of ESynC Algorithm

Step1 needs Time = $O(n)$ and Space = $O(n)$.

In Step2, constructing the $\delta$ near neighbor point sets for all points if using a simple method needs Time = $O(dn^2)$ and Space = $O(nd)$. In constructing the $\delta$ near neighbor point sets, time cost can be decreased by using the strategy of "space exchanges time" [2].

Step3 needs Time = $O(n)$ and Space = $O(n)$.

According to [1] and our analysis, ESynC algorithm needs Time = $O(Tdn^2)$, which is the same as SynC Algorithm. Here $T$ is the times of the while repetition in Step2.

### 4.3 Setting Parameter in ESynC Algorithm

Parameter $\delta$ can affect the results of clusters. In [1], parameter $\delta$ is optimized by the MDL principle [3]. In [19], two other methods were presented to estimate parameter $\delta$. Here, we can also set a proper parameter $\delta$ accordint to Theorem 1and Property 1.

### 4.4 The Convergence of ESynC Algorithm

In all simulations, if using ESynC algorithm to analyze the original data set $S$, then $S(T) = \{X_1(T), X_2(T), \ldots, X_n(T)\}$ will stay on some locations steadily after several iterations (many simulations only need 4 - 5 times). In the convergent result set $S(T)$, those locations that represent some points can be regarded as their cluster centers, and some locations that represent only one or several points are regarded as the final synchronization locations of isolaters.

The renewal computing of ESynC algorithm can be represented by a matrix formula, $\boldsymbol{S}(t + 1) = \boldsymbol{A}(t) \bullet \boldsymbol{S}(t)$. Here, $\boldsymbol{S}(t) = (X_1(t), X_2(t), \ldots, X_n(t))^T$ is the $t$-step location vector of $n$ points, $\{X_1, X_2, \ldots, X_n\}$, and $\boldsymbol{A}(t)$ is an $n \times n$ matrix. Suppose $\boldsymbol{S}(t = 0)$ is arranged by the final clusters order, which means that those points in the same clusters is coterminous in $\boldsymbol{S}(t = 0)$. In this case, $\boldsymbol{A}(T)$ will be a block matrix. For example, suppose the data set $S$ has $K$ final clusters or isolates, $|c_k|$ labels the number of points in the $k$ cluster or isolates, then the block matrix

$$A(T)=\begin{pmatrix}
1/|c_1| & \cdots & 1/|c_1| & & & & & & \\
\cdots & \cdots & \cdots & & 0 & & \cdots & & 0 \\
1/|c_1| & \cdots & 1/|c_1| & & & & & & \\
 & & & 1/|c_2| & \cdots & 1/|c_2| & & & \\
 & 0 & & \cdots & \cdots & \cdots & 0 & & 0 \\
 & & & 1/|c_2| & \cdots & 1/|c_2| & & & \\
 & 0 & & & 0 & & \cdots & & 0 \\
 & & & & & & 1/|c_K| & \cdots & 1/|c_K| \\
 & 0 & & & 0 & & \cdots & \cdots & \cdots \\
 & & & & & & 1/|c_K| & \cdots & 1/|c_K|
\end{pmatrix}$$ is a classical example.

Perhaps analyzing the synchronization process based on the Linear Vicsek Model from theory is more complex than analyzing Mean Shift Clustering algorithm [26] or multi-agent systems based on Vicsek Model [2, 3]. Here, we only give some results based on many simulations.

### 4.5 The improvements of ESynC Algorithm

One Improved version of ESynC algorithm (IESynC) can be obtained by combining multidimensional grid partitioning method and Red-Black tree structure to constructe the near neighbor point sets of all points. The improving method that can decrease its time cost is introduced detailly in [17]. Generally, we first partition the data space of the data set $S = \{X_1, X_2, \ldots, X_n\}$ by using a kind of multidimensional grid partitioning method. Then construct an effective index of all grid cells and compute $\delta$ near neighbor grid cell set of each grid cell. If every grid cell uses a Red-Black tree to index its data points in each synchronization step, then constructing the $\delta$ near neighbor point set of every point will become quicker when the number of grid cells is proper.

Another improvement in both time cost and space cost of ESynC algorithm is described in the next subsection.

## 5. A Shrinking Synchronization Clustering Algorithm Based on Another Linear Vicsek Model

This method has similar process with SynC algorithm [1] and ESynC algorithm except using a different dynamics clustering model. The synchronization model represented by Eq.(12) can be used for clustering a core set.

### 5.1 The Description of SSynC Algorithm

**Algorithm Name**: a Shrinking Synchronization Clustering algorithm (SSynC)

**Input**: data set $S = \{X_1, X_2, \ldots, X_n\}$, dissimilarity measure $d(\cdot, \cdot)$, parameter $\delta$, and parameter $\varepsilon$ (a very small real).

**Output**: The final core set $C(T) = \{C_1(T), C_2(T), \ldots, C_n(T)\}$.

**Procedure**:

Step1. Initialization:

```
IterateStep t is set as zero firstly, that is: t ← 0;
* Create initial core set C(t = 0). */
for (i = 1; i ≤ n; i++)
{
    Ci(t = 0).Core_Id ← i;
    Ci(t = 0).Core_Location ← Xi;
    Ci(t = 0).Parent_CoreId ← i;
    Ci(t = 0).Number_ContainingPoints ← 1;
}
* Create initial active point set AP(t = 0). */
AP(t = 0) ← {X1, X2, …, Xn};
NumberOfAP(t = 0) ← n;
```

Step2. Execute the iterative synchronization process of the dynamical clustering:

```
while ((the dynamical clustering does not satisfy its convergent condition) and (t < 50))
{
    for (each point Y(t) in the active point set AP(t))
    {
        Construct the δ near neighbor point set δ(Y(t+1)) of point Y(t+1) according to Definition 1;
        Compute the renewal value, Y(t+1), of Y(t) using Eq.(12);
    }
    for (each unlabelled point Y(t+1) in the point set AP(t+1))
    {
        the number "Core_Location" of the correspoing core point of point Y(t+1) is updated by the value of Y(t+1);
        Construct the ε near neighbor point set ε(Y(t+1)) of point Y(t+1) according to Definition 1;
        for (each unlabelled point Z(t+1) in the ε near neighbor point set ε(Y(t+1)) of point Y(t+1))
        {
            point Z(t+1) is labelled as unactive point;
            the number "Parent_CoreId" of the correspoing core point of point Z(t+1) is assigned by the number "Core_Id" of the correspoing core point of point Y(t+1);
            the number "Number_ContainingPoints" of the correspoing core point of point Z(t+1) is added into the number "Number_ContainingPoints" of the correspoing core point of point Y(t+1);
        }
    }
    Delete all labelled unactive points from AP(t+1);
    Compute the number of unlabelled points of the renewal active point set AP(t+1) to NumberOfAP(t+1);
    IterateStep t is increased by 1, that is: t++;
    if (NumberOfAP(t+1) == NumberOfAP(t))    /* The number of points in the renewal active point set AP(t+1) is equal to the number of points in the active point set AP(t) */
        We think the dynamical clustering reaches its convergent result, and then exit from the while repetition;
}
Compress the paths of some unactive core points in the C(t) just like the joint-set
```

method.

Step3. Finally we get a core set $C(T) = \{C_1(T), C_2(T), \ldots, C_n(T)\}$, where $T$ is the times of the while repetition in Step2. The final set $C(T)$ reflects the natural clusters or isolate points of the data set $S$.

For example, if $C_i(T)$.Core_Id == $C_i(T)$.Parent_CoreId and $C_i(T)$.Number_ContainingPoints == 1 are satisfied, we can think the $i$-th point is an isolate; if $C_i(T)$.Core_Id == $C_i(T)$.Parent_CoreId and $C_i(T)$.Number_ContainingPoints >> 1 are satisfied, we can think the $i$-th point is a cluster core that represents some other points.

**5.2 Time Complexity Analysis of SSynC Algorithm**

Step1 needs Time = $O(n)$ and Space = $O(n)$.

In the first synchronization process of Step2, constructing the $\delta$ near neighbor point sets for all points if using a simple method needs Time = $O(dn^2)$ and Space = $O(nd)$. In the $t$-step synchronization process of Step2, constructing the $\delta$ near neighbor point sets for all points if using a simple method needs Time = $O(dn_{(t)}^2)$ and Space = $O(n_{(t)}d)$, where $n_{(t)}$ is the number of active cores in the $t$-step synchronization process. In constructing the $\delta$ near neighbor point sets, time cost can be decreased by using the strategy of "space exchanges time" [2].

Step3 needs Time = $O(n)$ and Space = $O(n)$.

According to [1] and our analysis, ESynC algorithm needs Time = $O(d\cdot(n_{(t=0)}^2 + n_{(t=1)}^2) + \ldots + n_{(t=T-1)}^2)) < O(Tdn^2)$, which is usually less than SynC algorithm. Here $T$ is the times of the while repetition in Step2.

**5.3 Setting Parameters in SSynC Algorithm**

Parameter $\delta$ that affects the results of clusters is the same as ESynC algorithm. In [1], parameter $\delta$ is optimized by the MDL principle [3]. In [19], two other methods were presented to estimate parameter $\delta$. Here, we can also set a proper parameter $\delta$ accordint to Theorem 1and Property 1.

Parameter $\varepsilon$ can affect the time cost of SSynC Algorithm. In simulations, we get the same clustering results except time cost for two different values (0.00001 and 1) of parameter $\varepsilon$. Usually, the larger parameter $\varepsilon$ is set, the more time cost of SSynC algorithm needs.

Figure 3 describes the number of active cores with time evolution using SSynC algorithm. From Fig. 3, we observe that parameter $\delta$ can affect the number of active cores with time evolution, which affect the time cost of SSynC algorithm.

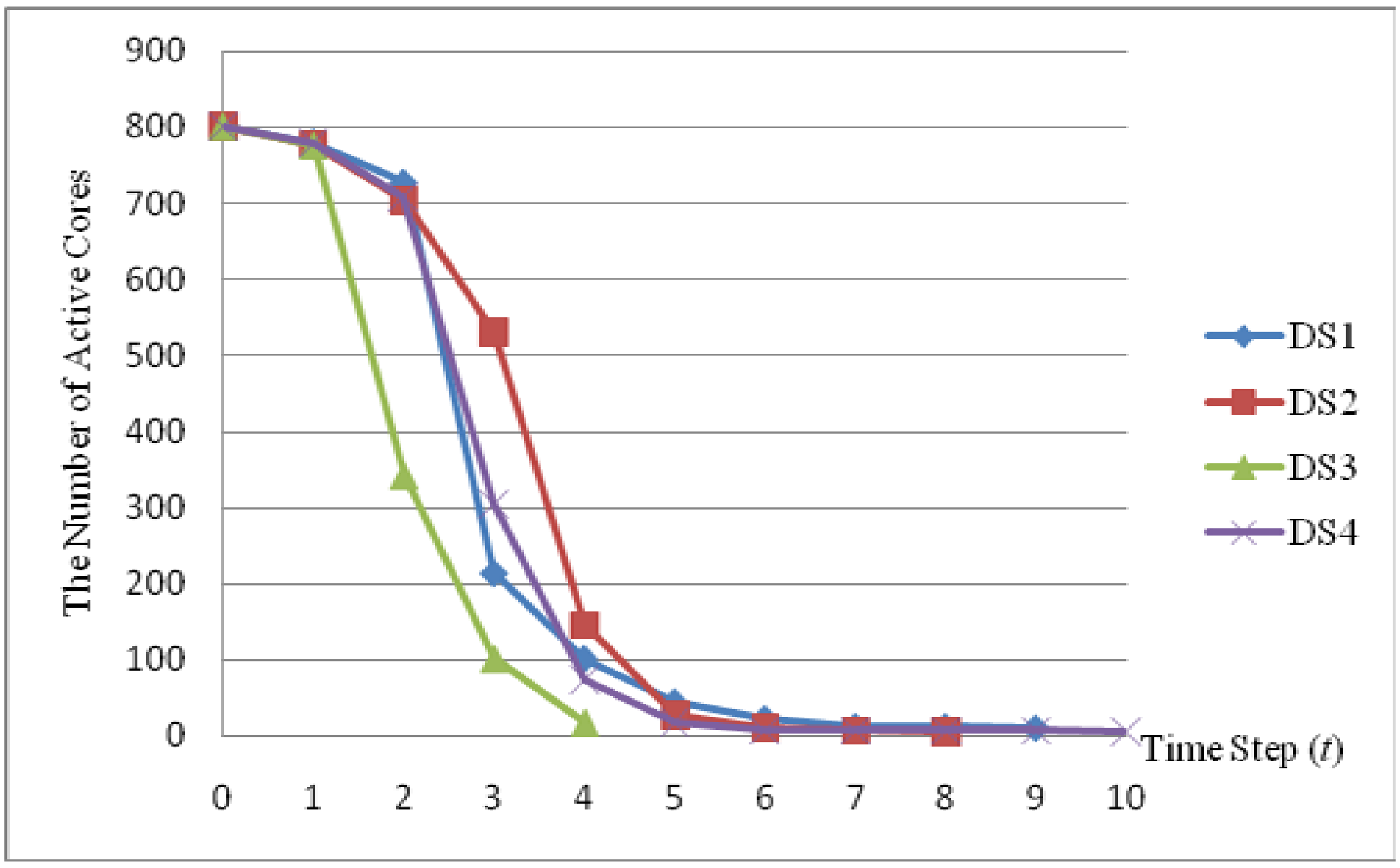

(a) parameter $\varepsilon = 0.00001$

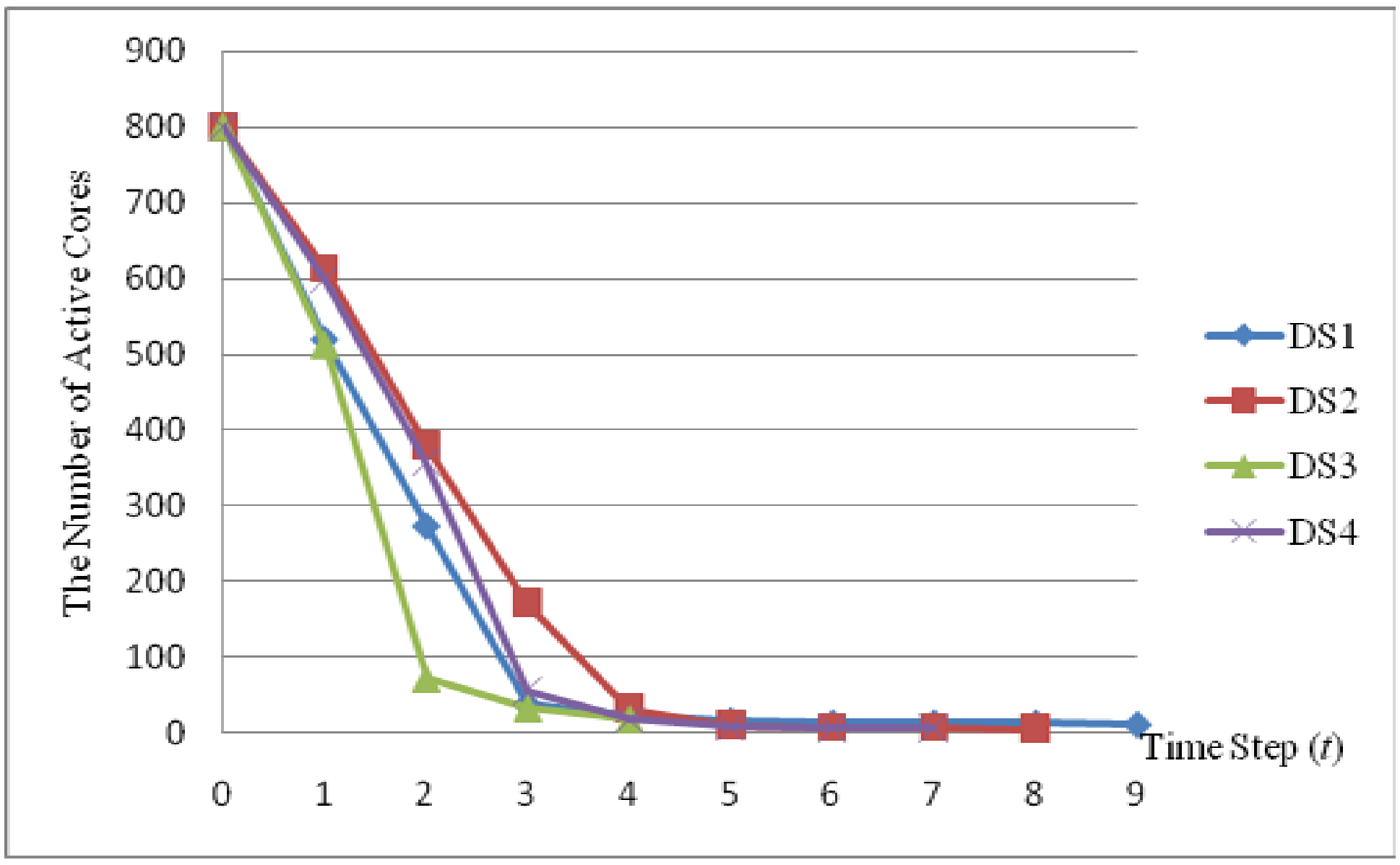

(b) parameter $\varepsilon = 1$

Figure 3. The Number of Active Cores with Time Evolution using SSynC Algorithm. In Fig. 3, parameter $\delta$ is set as 18, parameter $\varepsilon$ (a very small real) is set as 0.00001 or 1, and the number of points in the four data sets is all set as 800.

**5.4 The improvement of SSynC Algorithm**

One Improved version of SSynC algorithm can be obtained by combining multidimensional grid partitioning method and Red-Black tree structure to constructe the near neighbor point sets of all active cores. The improving method that can decrease its time cost is introduced detailly in [17]. Generally, we first partition the data space of the data set $S = \{X_1, X_2, \ldots, X_n\}$ by using a kind of multidimensional grid partitioning method. Then construct an effective index of all grid cells and compute $\delta$ near neighbor grid cell set of each grid cell. If every grid cell uses a Red-Black tree to index its active cores in each synchronization step, then constructing the $\delta$ near neighbor point set of every active core will become quicker when the number of grid cells is proper.

Before iterative evolution, if we set a proper value of paremater $\delta$ to filtrate isolates, then these isolates can be set as no active cores that will not be operated in the next iterative evolutions. This improvement of implementation technique is often effective for some data sets.

# 6. An Effective Multi-level Synchronization Clustering Algorithm Facing to Big Data

Facing to big data, usual clustering method can not explore all data in memory for one time. In order to conquer this problem, we present an effective Multi-level Synchronization Clustering algorithm (MSynC) by using an extended linearized version of Vicsek model.

Figure 4 presents the framework of MsynC algorithm.

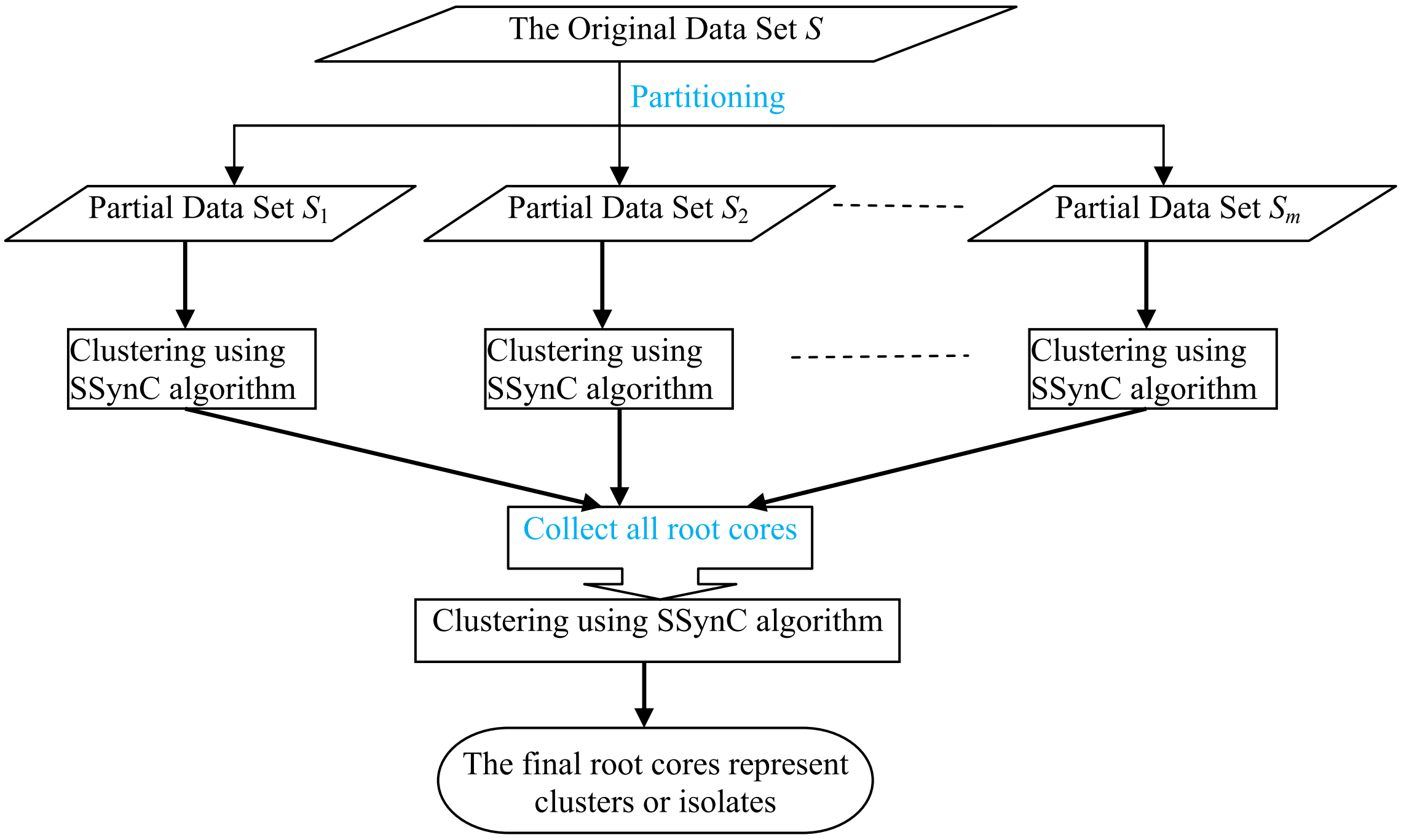

Figure 4. The Framework of MsynC algorithm

## 6.1 The Partitioning of Big Data in MSynC Algorithm

We can use a random partitioning method to segment a big data set. If the partitioning is uniform, then MSynC algorithm will get similar results with SSynC algorithm.

### 6.2 The Description of MSynC Algorithm

**Algorithm Name**: an effective Multi-level Synchronization Clustering algorithm (MSynC)

**Input**: data set $S = \{X_1, X_2, \ldots, X_n\}$, dissimilarity measure $d(\cdot, \cdot)$, parameter $\delta$, parameter $m$, and parameter $\varepsilon$ (a very small real).

**Output**: The final core set $C(T) = \{C_1(T), C_2(T), \ldots, C_n(T)\}$.

**Procedure**:

Step1. Partitioning the data set $S = \{X_1, X_2, \ldots, X_n\}$ that readed from a file or a data base into $m$ subsections, $\{S_1, S_2, \ldots, S_m\}$. Usually, we need $S = S_1 \cup S_2 \cup \ldots \cup S_m$.

Step2. Clustering each part from $\{S_1, S_2, \ldots, S_m\}$ using SSynC algorithm:

```
for (i = 1; i ≤ m; i++)
    SSynC(subset Si, parameter δ, parameter ε);
```

Step3. Creat a new set, *CS*, by collecting all root cores from the clustering results of $\{S_1, S_2, \ldots, S_m\}$.

Step4. Clustering the collected core set *CS* using SSynC algorithm.

Step5. The final root cores of the collected core set *CS* represent clusters or isolates of the data set $S = \{X_1, X_2, \ldots, X_n\}$.

### 6.3 Time Complexity Analysis of MSynC Algorithm

Step1 needs Time = $O(n)$ and Space = $O(n)$.

Step2 needs Time = $O(d\cdot(n_{(t=0)}^2 + n_{(t=1)}^2) + \ldots + n_{(t=T-1)}^2)/m)$.

Suppose subsection $S_i$ ($i = 1, 2, \ldots, m$) has $K_i$ clusters, then Step3 needs $O(K_1 + K_2 + \ldots + K_m) = O(Km)$.

Step4 needs Time = $O(d\cdot((Km_{(t=0)})^2 + (Km_{(t=2)})^2 + \ldots + (Km_{(t=T-1)})^2))$

According to our analysis, usually MSynC algorithm needs less time than SSynC algorithm.

### 6.4 Setting Parameters in MSynC Algorithm

Parameter $\delta$ that affects the results of clusters is the same as ESynC algorithm. In [1], parameter $\delta$ is optimized by the MDL principle [3]. In [19], two other methods were presented to estimate parameter $\delta$. Here, we can also set a proper parameter $\delta$ accordint to Theorem 1and Property 1.

Parameter $\varepsilon$ can affect the time cost of SSynC Algorithm. In simulations, we get the same clustering results except time cost for two different values (0.00001 and 1) of parameter $\varepsilon$ is set as. Usually, the larger parameter $\varepsilon$ is set, the more time cost of SSynC algorithm needs.

Parameter $m$ can affect the time cost of MSynC algorithm. Usually, the larger parameter $m$ is set, the less computing time cost and the more communication time cost SSynC algorithm needs.

# 7. Simulated Experiments

## 7.1 Experimental Design

Our experiments are finished in a personal computer (Capability Parameters: Pentium(R) Dual CPU T3200 2.0GHz, 2G Memory). Experimental programs are developed using Visual C++6.0 under Windows XP.

To verify the improvements in time complexity of our algorithms, there will be some experiments of some artificial data sets in the next subsections.

Four kinds of artificial data sets (DS1 – DS4) are produced in a 2-D region [0, 600] × [0, 600] by a program. Eight kinds of artificial data sets (DS5 – DS12) are produced in a range [0, 600] in each dimension by a similar program. Table 1 is the description of the artificial data sets.

Table 1. The description of the artificial data sets

| Data Sets (*DS*) | Number of Clusters (*NC*) | With Noise | Cluster Semidiameter (*CS*) | Dimension (*d*) |
|---|---|---|---|---|
| DS1 | 5 | yes | 40 | 2 |
| DS2 | 5 | no | 50 | 2 |
| DS3 | 9 | yes | 30 | 2 |
| DS4 | 9 | no | 40 | 2 |
| DS5 | 12 | no | 30 | 2 |
| DS6 | 12 | no | 30 | 4 |
| DS7 | 12 | no | 30 | 6 |
| DS8 | 12 | no | 30 | 8 |
| DS9 | 12 | no | 30 | 10 |
| DS10 | 12 | no | 30 | 12 |
| DS11 | 12 | no | 30 | 14 |
| DS12 | 12 | no | 30 | 16 |

In our simulated experiments, the maximum times of synchronization moving in the while repetition of SynC algorithm, ESynC algorithm, IESynC algorithm, and SSynC algorithm is set as 50.

Comparative results of these algorithms are presented by some figures (Figure 5 - Figure 9) and some tables (Table 2 - Table 3). And performance of algorithms is measured by time cost (second).

Subsection 7.2 and subsection 7.3 present some experimental results of some artificial data sets.

In the experiments, parameter $r_i$ ($i$ = 1, 2, …, $d$) is the interval length in the $i$-th dimension of grid cell [18], and $\delta$ is the threshold parameter in Definition 1. The detailed discussion on how to construct grid cells and $\delta$ near neighbor points is described in [18]. How to select a proper parameter $\delta$ for SynC algorithm is discussed detailly in [1]. ESynC algorithm can use the same method as SynC algorithm to select a proper parameter $\delta$. In ESynC algorithm, setting different values of parameter $r_i$ ($i$ = 1, 2, …, $d$) for different dimensions will result in different number of grid cells and different time cost.

In DBSCAN algorithm, paprmaeter*MinPts* = 4 and paprmaeter Eps is equal to our parameter $\delta$ in the same case.

**7.2 Experimental Results of Some Artificial Data Sets (DS1 – DS4)**

7.2.1 Comparison results among SynC algorithm, ESynC algorithm, IESynC algorithm, and SSynC algorithm

Table 2. Comparison of several different synchronization algorithms by using four kinds of artificial data sets (DS1 – DS4). In Tab 3, parameter $\delta$ = 18, the number of data points $n$ = 10000, parameter $\varepsilon$ (a very small real, if the distance of two points is less than $\varepsilon$, then they are regarded as in the same cluster; it also used as the iterative exit threash) = 0.00001. In IESynC, $r_1, r_2$ = 10.

| Comparision of Algorithms | | **DS1** | **DS2** | **DS3** | **DS4** |
|---|---|---|---|---|---|
| **Spend time (second)** | SynC | 448 | 553 | 538 | 525 |
| | ESynC | **56** | **70** | **107** | **81** |
| | IESynC | **56** | **66** | **75** | **55** |
| | SSynC | **52** | **69** | **34** | **52** |
| **Iterative times** | SynC | 41 | 50 | 50 | 50 |
| | EsynC, IESynC, SSynC | **4** | **5** | **8** | **6** |
| **The number of steady locations** | SynC | 254 | 379 | 260 | 431 |
| | EsynC, IESynC, SSynC | **14** | **5** | **25** | **8** |
| **The cluster order parameter $r_c$** | SynC | 42.7595 | 29.1156 | 44.8862 | 25.0379 |
| | EsynC, IESynC | **1995.40** | **1999.00** | **1352.31** | **1356.98** |
| ***AveLen*(*T*)** | SynC | 11.0132 | 11.1481 | 10.6701 | 11.0542 |
| | EsynC, IESynC | **0** | **0** | **0** | **0** |

**Note:** *AveLen*($T$): The final average length of edges in the final $\delta$ near neighbor undirected graph. Here, $T$ is equal to the iterative times of one synchronization

clustering algorithm. In Table 2, two indexes, the cluster order parameter $r_c$ and *AveLen*(*T*), are unfit for measuring the clustering effect of SSynC algorithm.

From Table 2, by intercomparing SynC, ESynC, IESynC, and SSynC, we find that SSynC is the fastest clustering algorithm.

7.2.2 Comparison results among SynC algorithm, ESynC algorithm, IESynC algorithm, SSynC algorithm, and some classical clustering algorithms

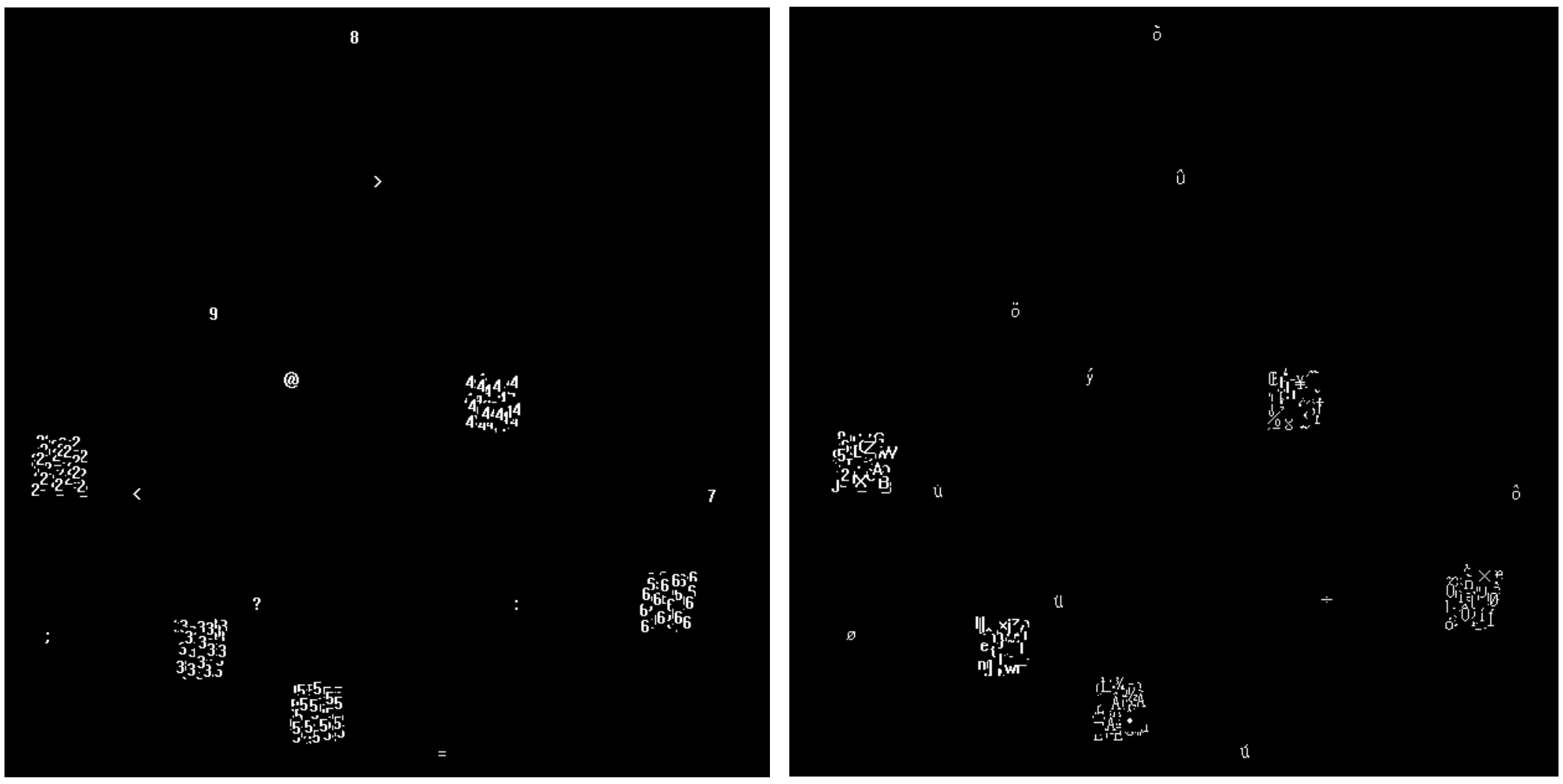

(a) Clusters identified by ESynC, IESynC, and SSynC (15 clusers or isolates) (b) Clusters identified by SynC (204 clusers or isolates)

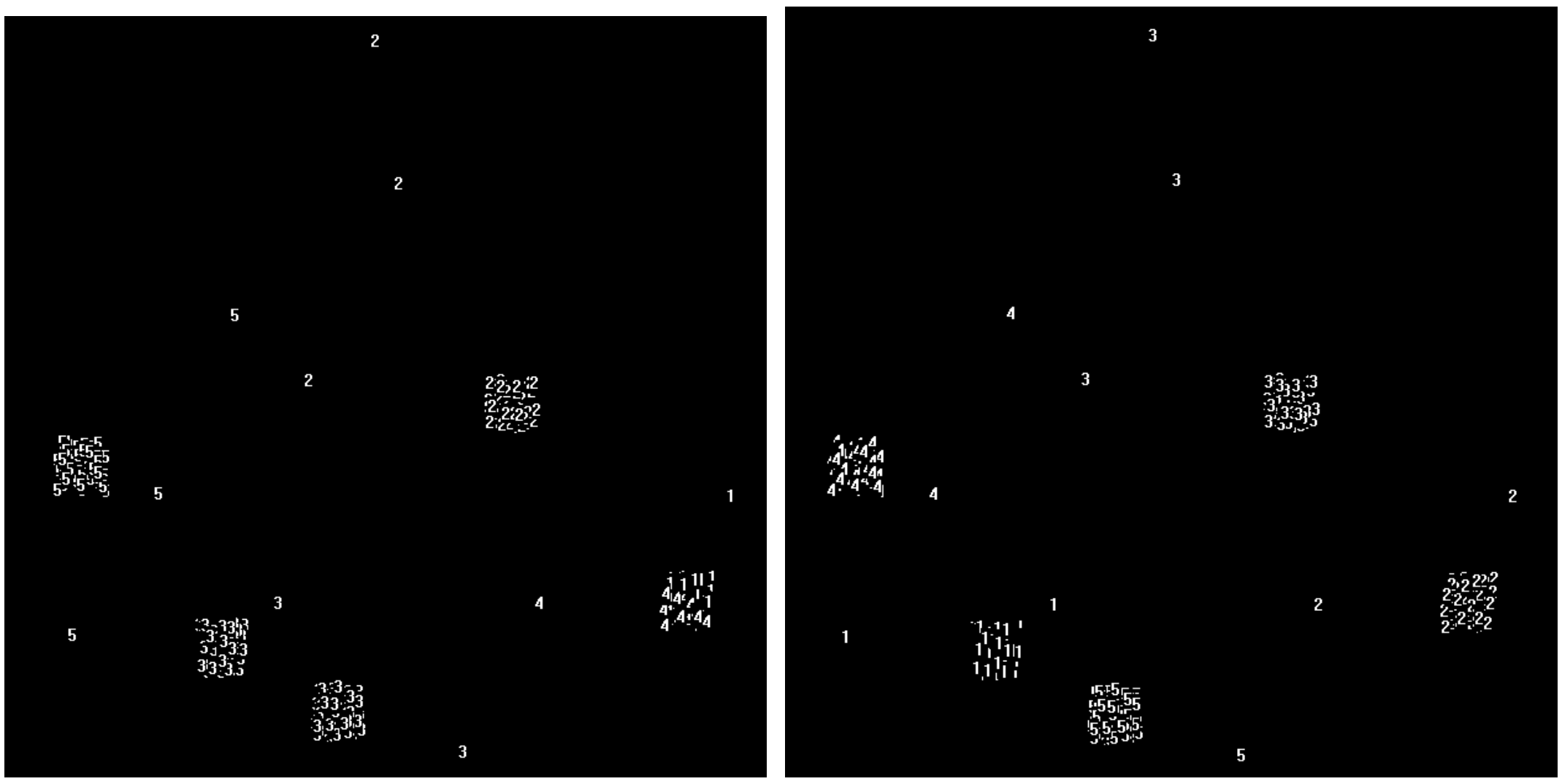

(c) Clusters identified by Kmeans (predefined 5 clusers) (d) Clusters identified by FCM (predefined 5 clusers)

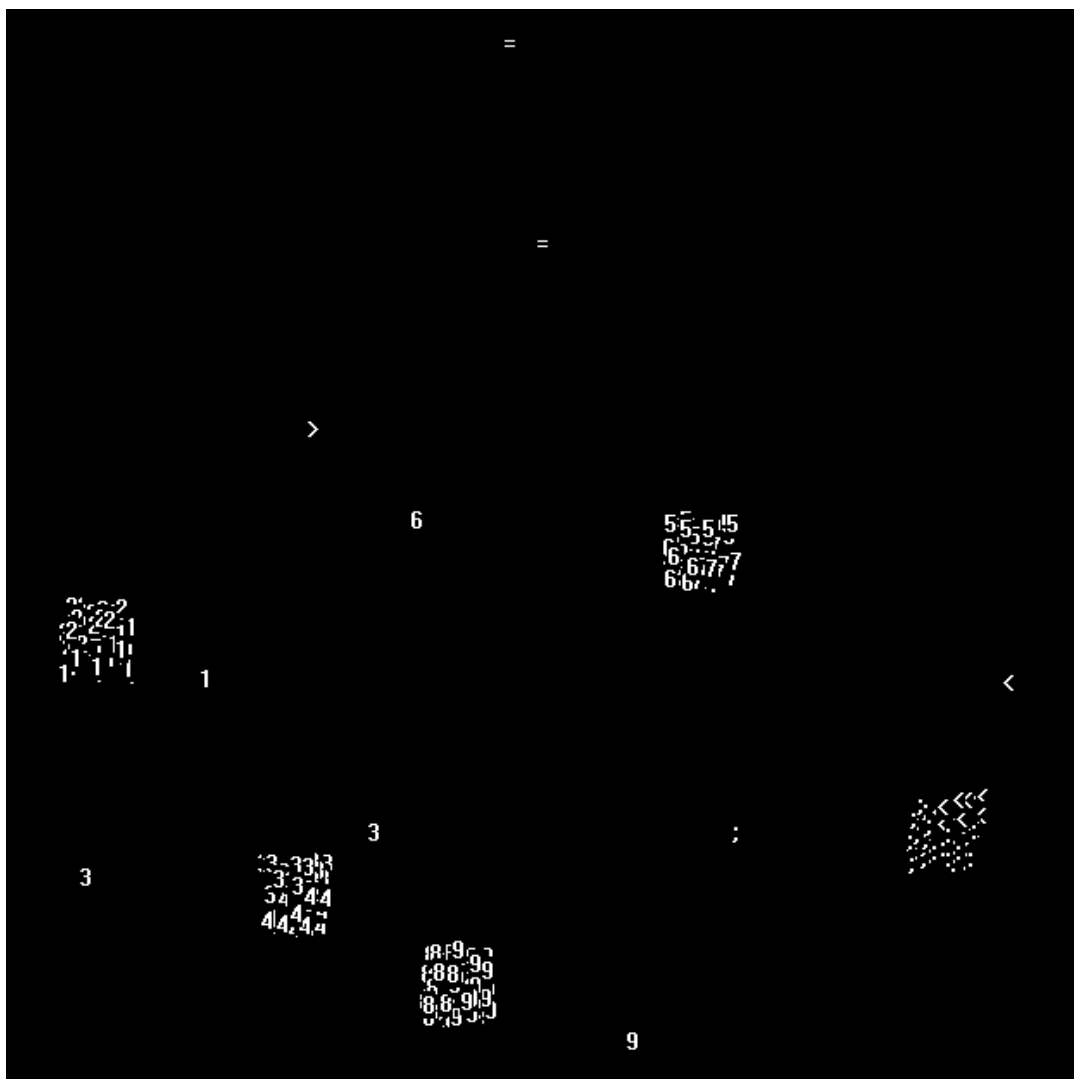

(e) Clusters identified by AP (14 clusers)

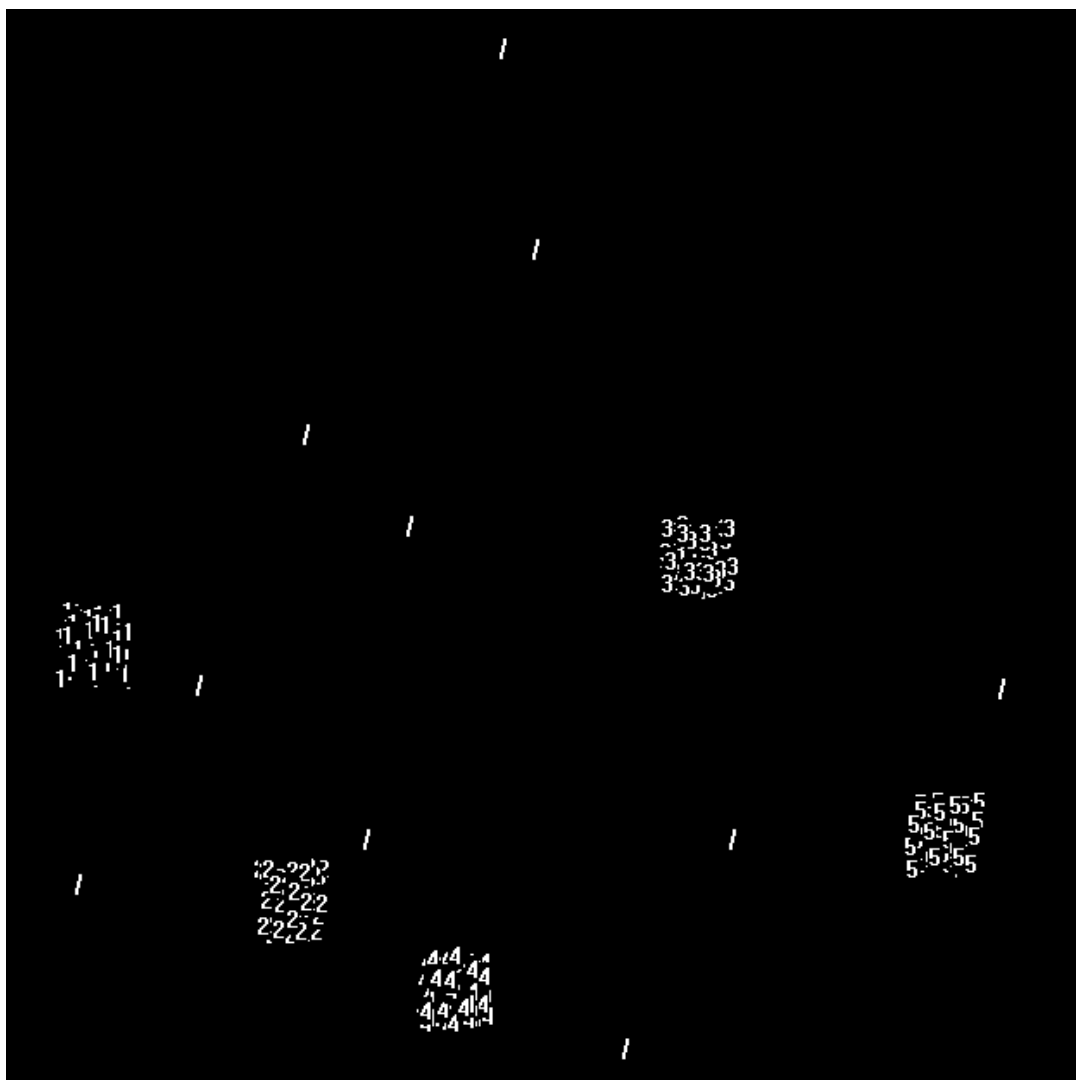

(f) Clusters identified by DBSCAN (5 clusers)

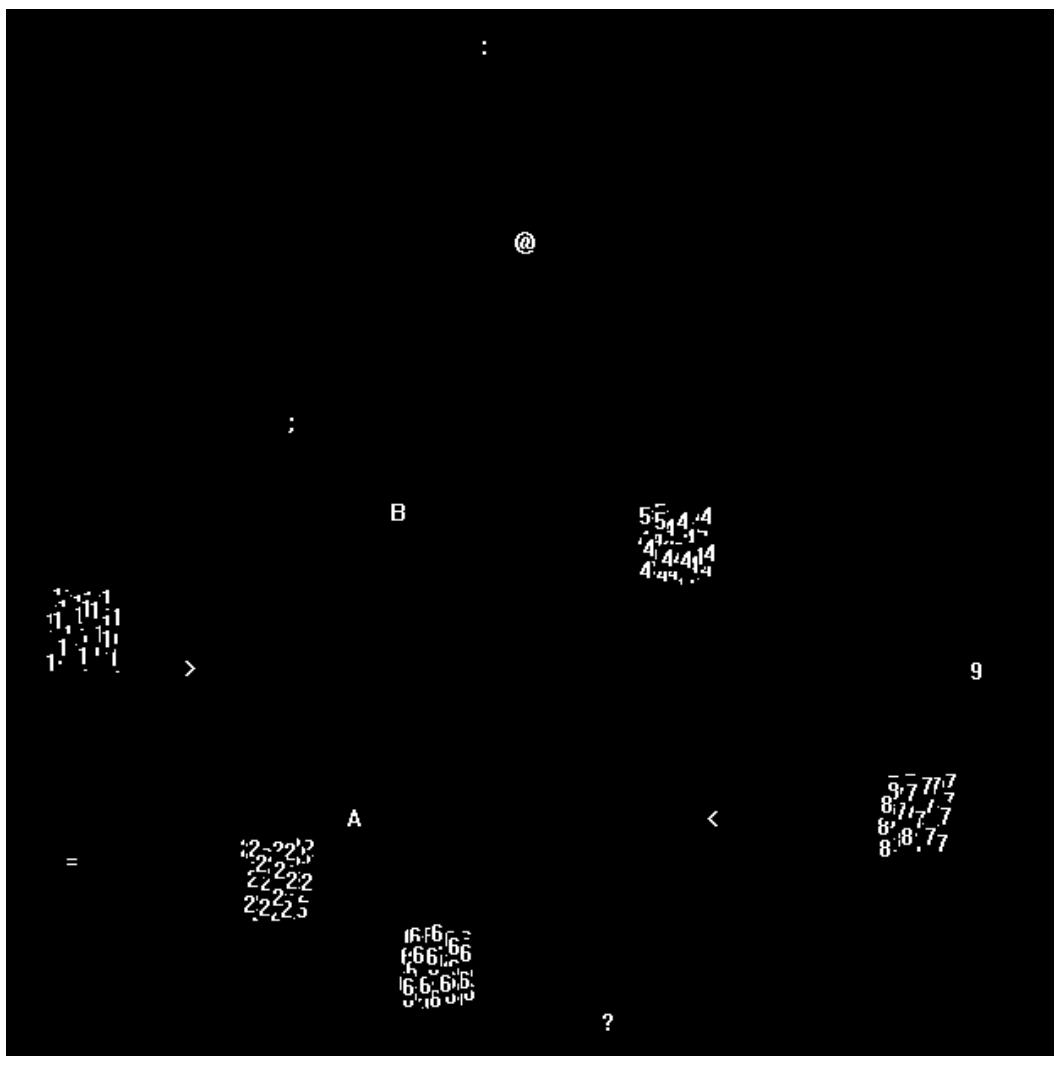

(g) Clusters identified by MeanShiftCluster (18 clusers)

Figure 5. Comparison of clustering results of several algorithms (DS1, $n$ = 400)

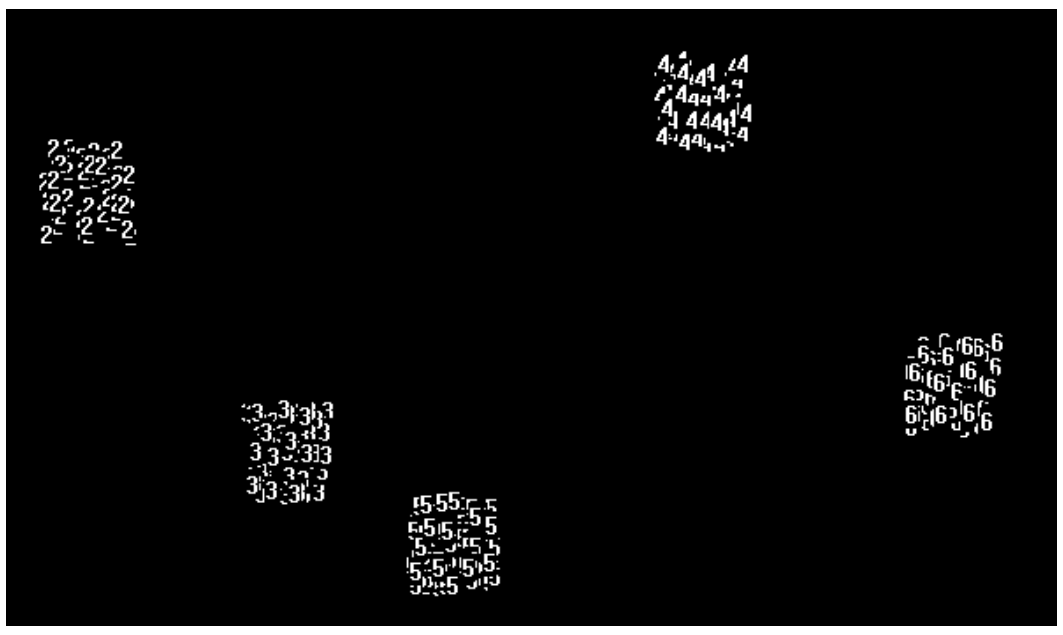

(a) Clusters identified by ESynC, IESynC, and SSynC (5 clusers)

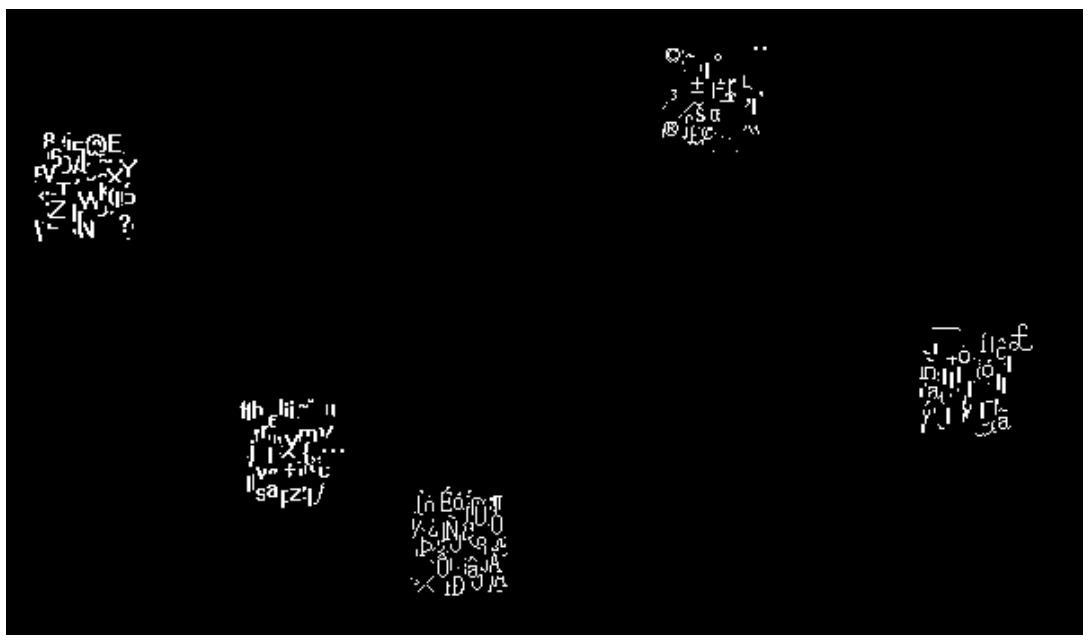

(b) Clusters identified by SynC (227 clusers)

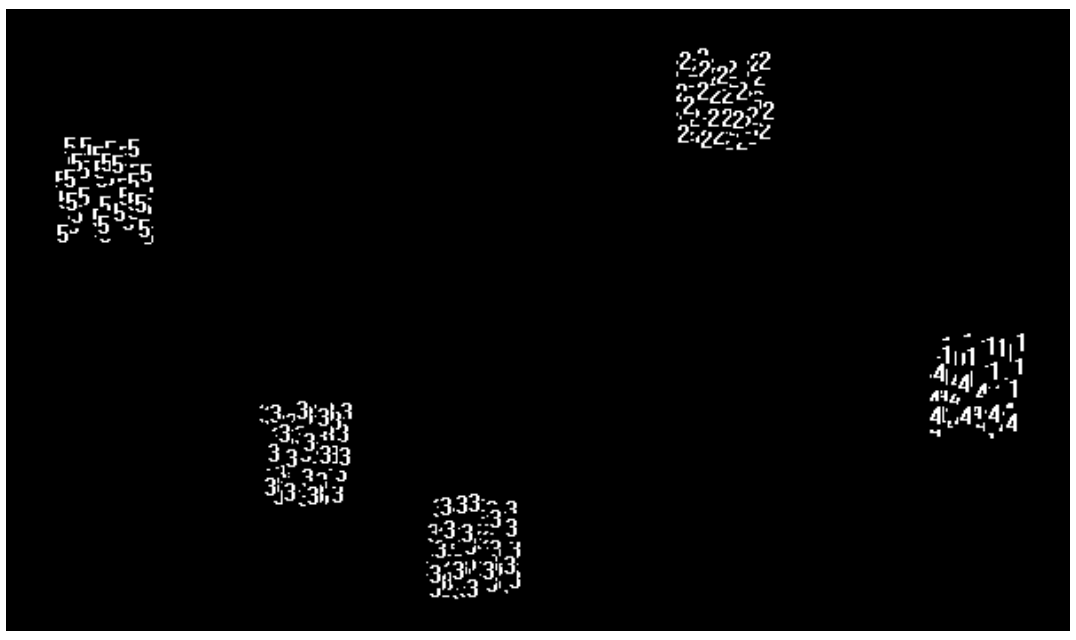

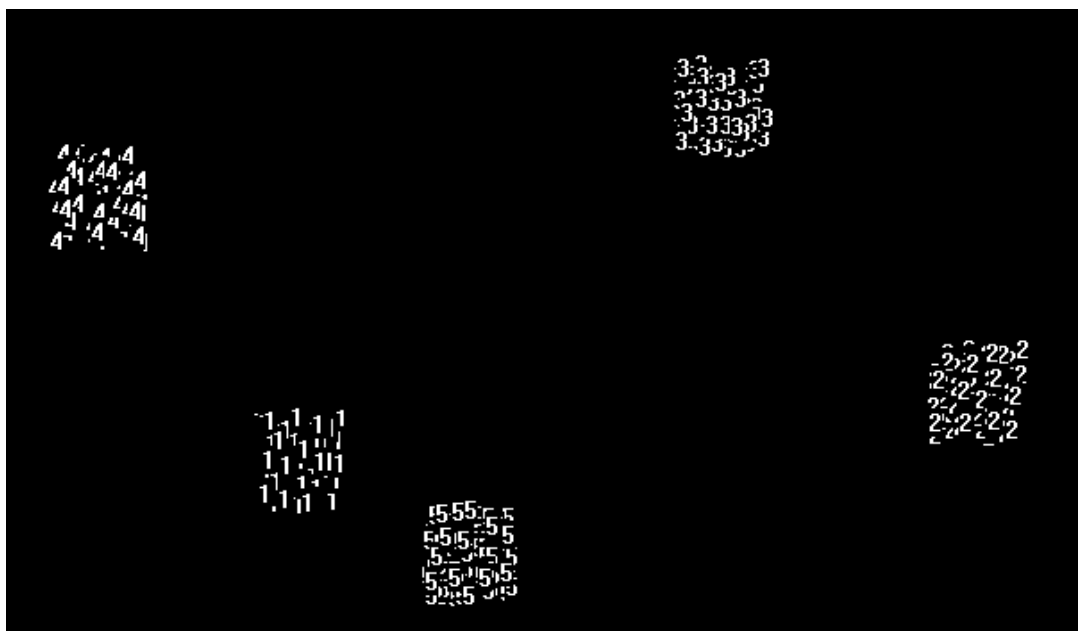

(c) Clusters identified by Kmeans (predefined 5 clusers) (d) Clusters identified by FCM (predefined 5 clusers)

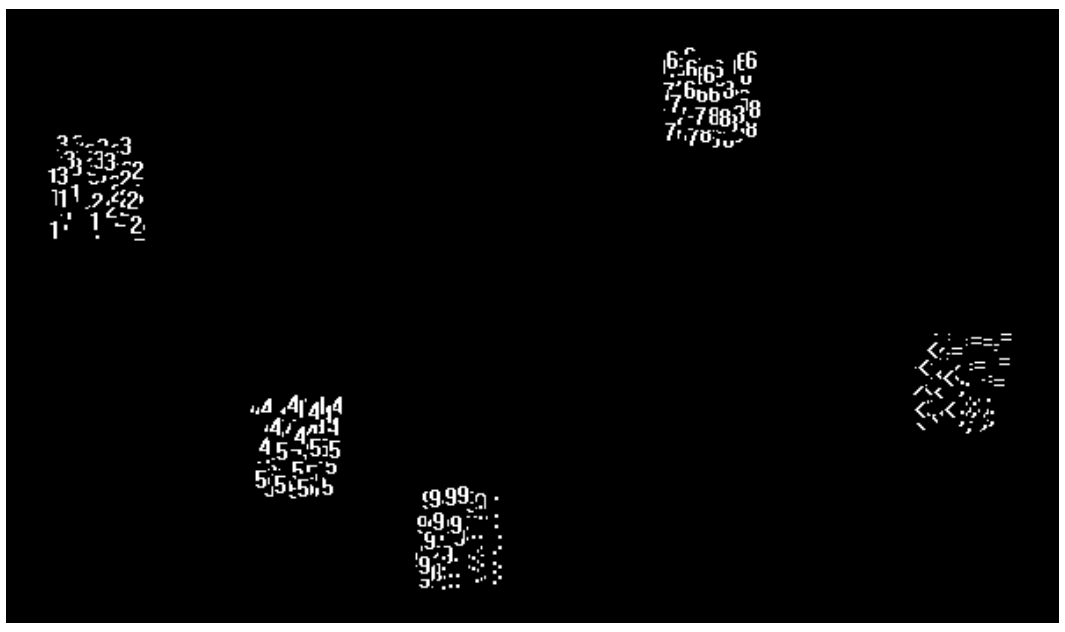

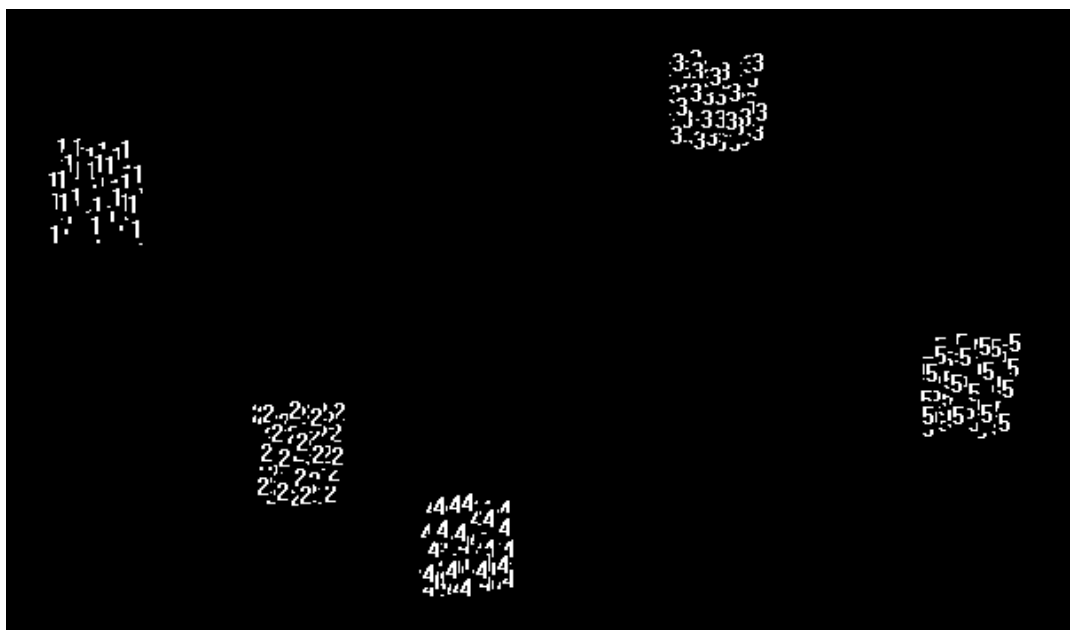

(e) Clusters identified by AP (13 clusers) (f) Clusters identified by DBSCAN (5 clusers)

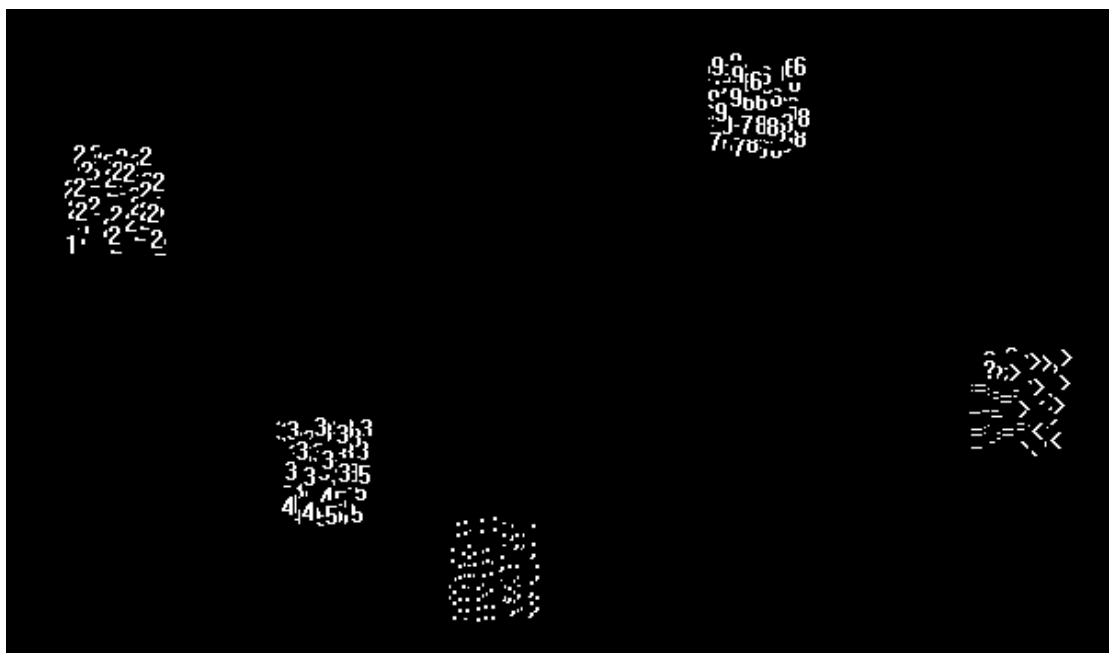

(g) Clusters identified by MeanShiftCluster (15 clusers)

Figure 6. Comparison of clustering results of several algorithms (DS2, $n$ = 400)

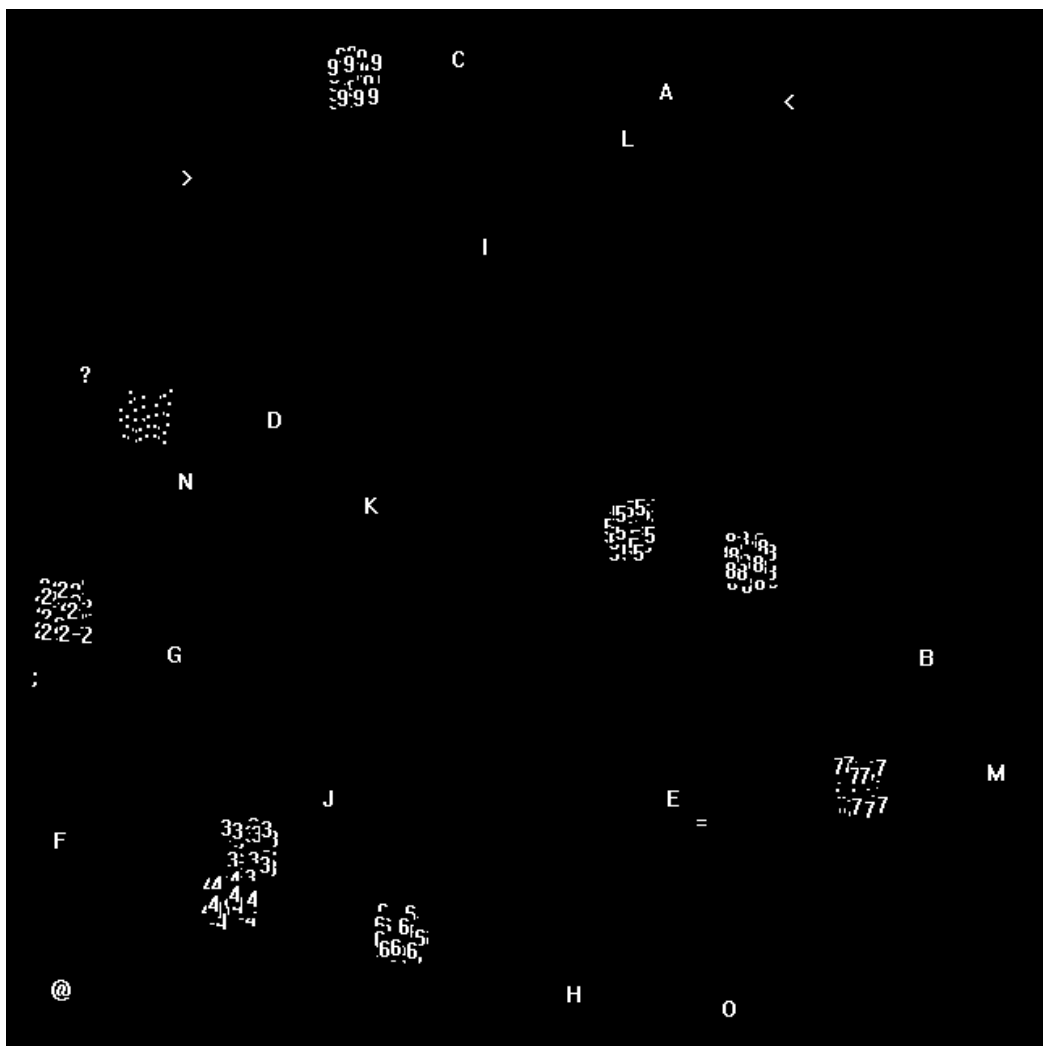

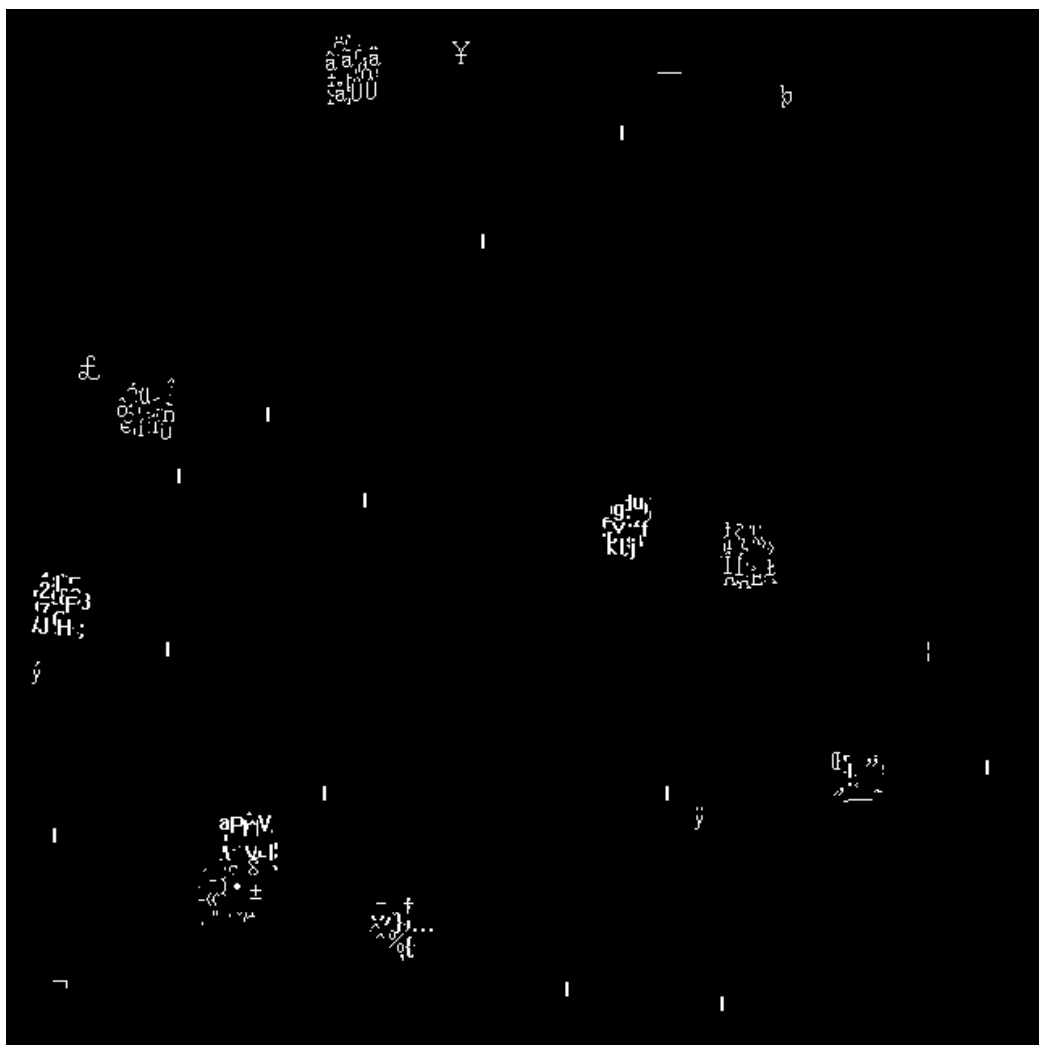

(a) Clusters identified by ESynC, IESynC, and SSynC (30 clusers or isolates) (b) Clusters identified by SynC (224 clusers or isolates)

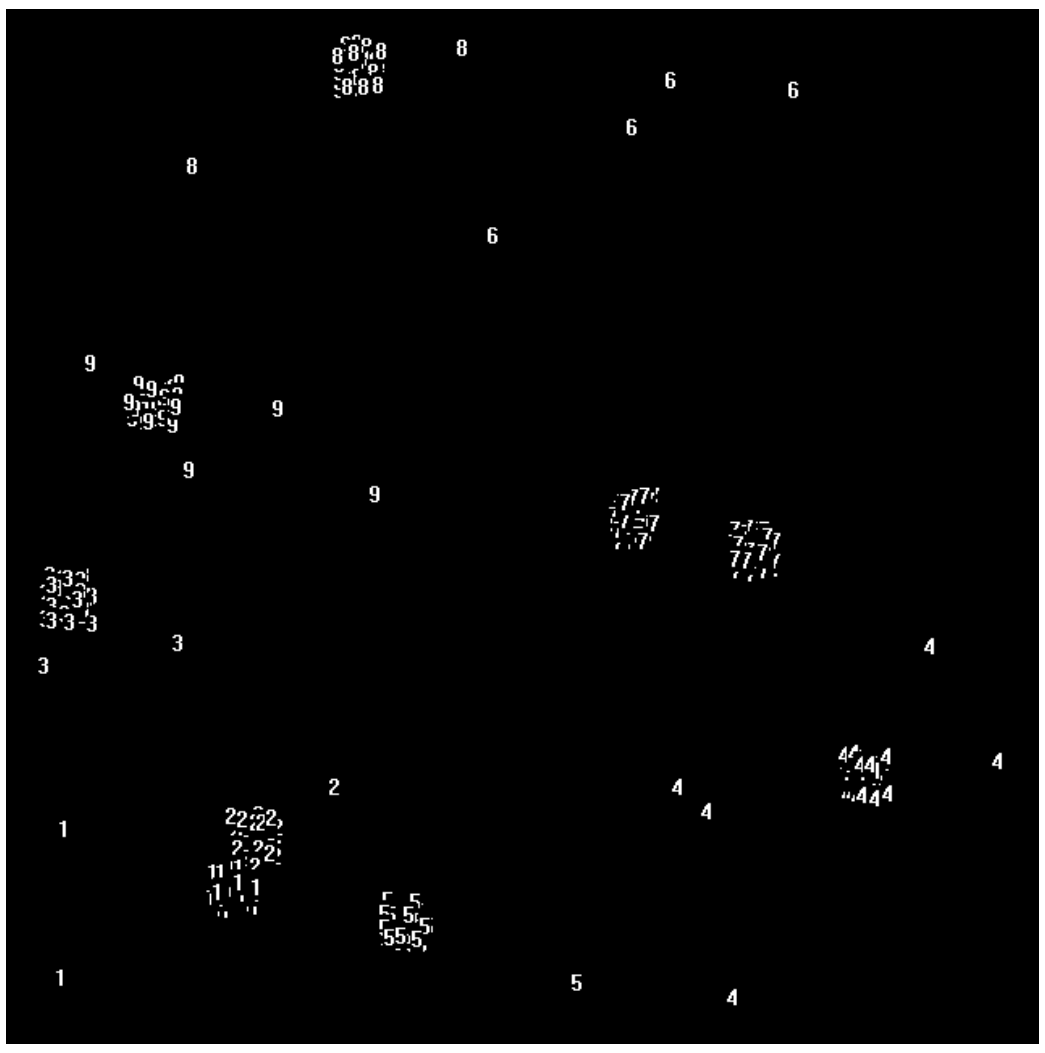

(c) Clusters identified by Kmeans (predefined 9 clusers)

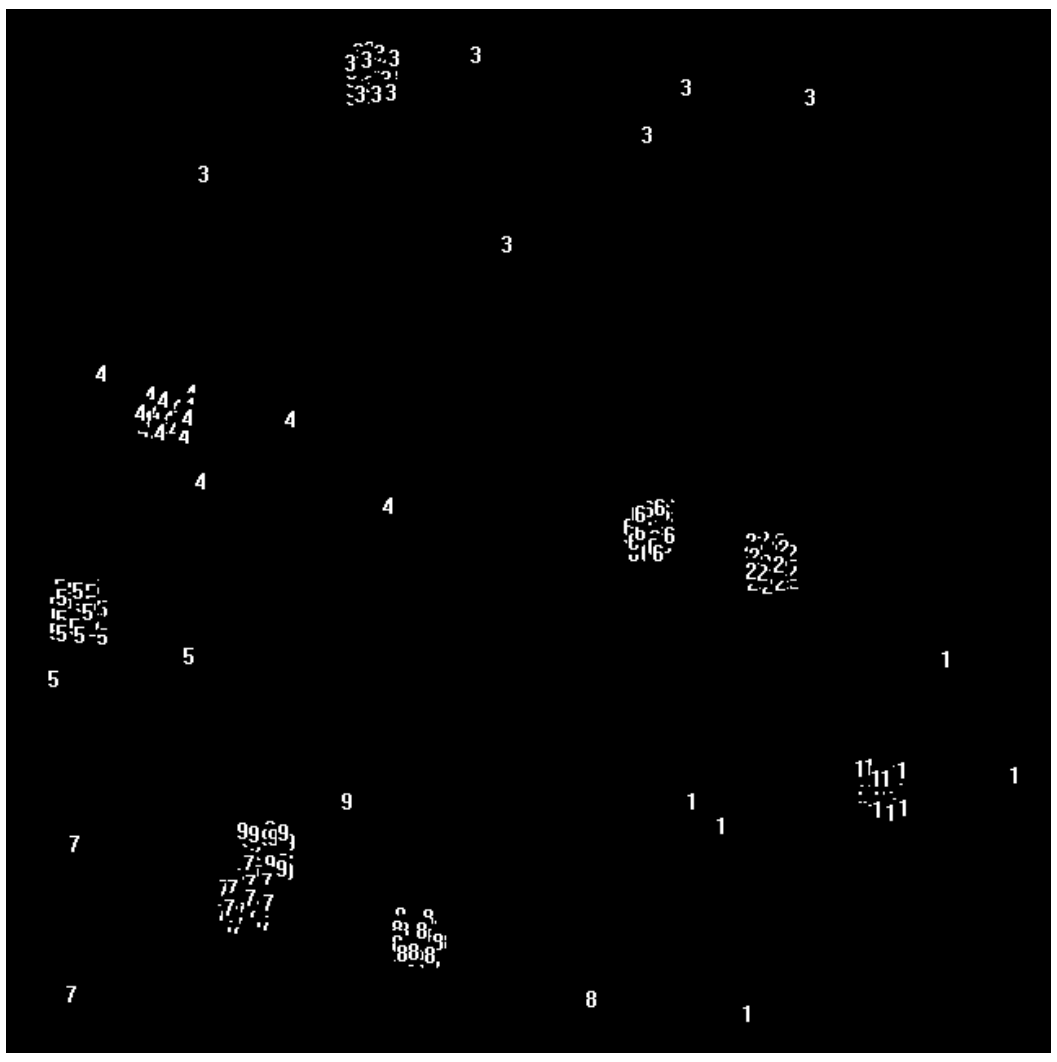

(d) Clusters identified by FCM (predefined 9 clusers)

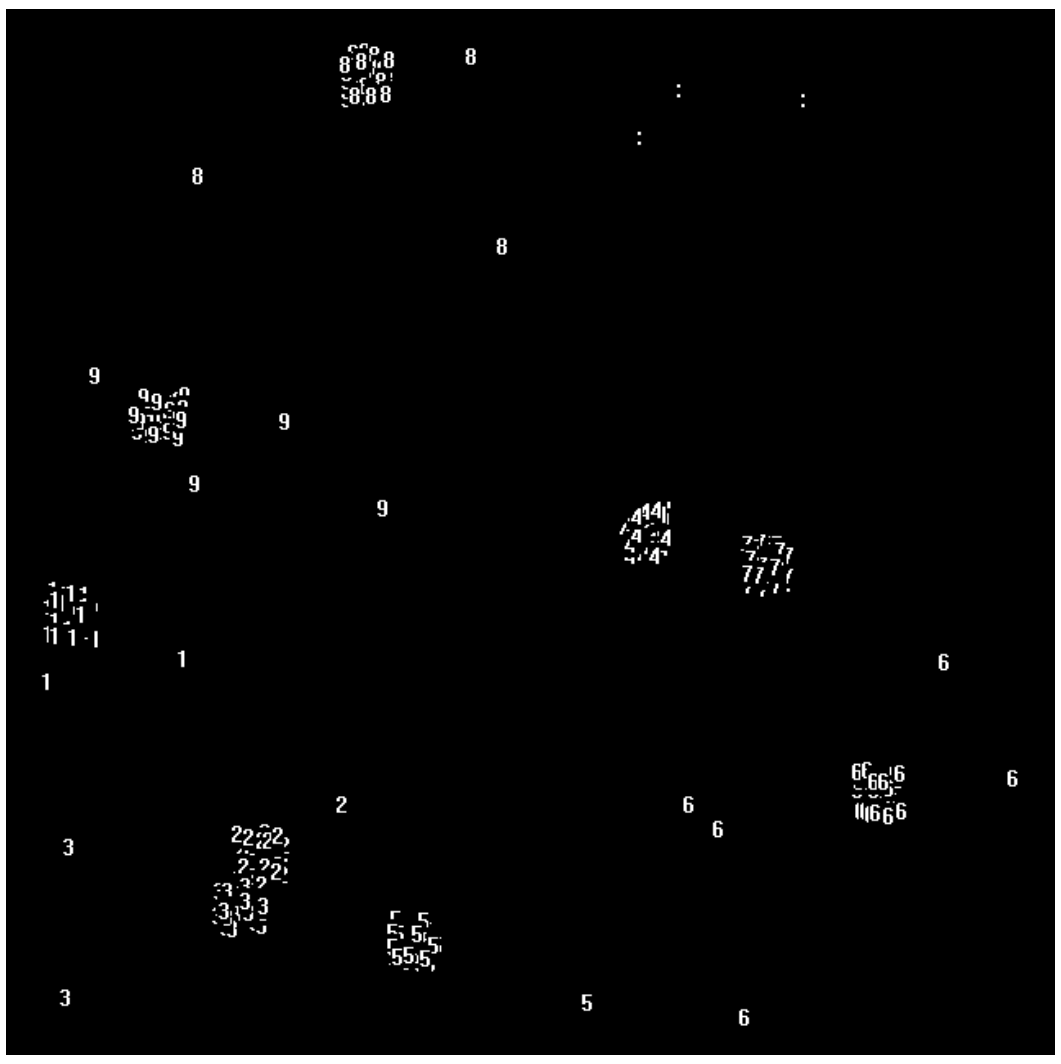

(e) Clusters identified by AP (10 clusers)

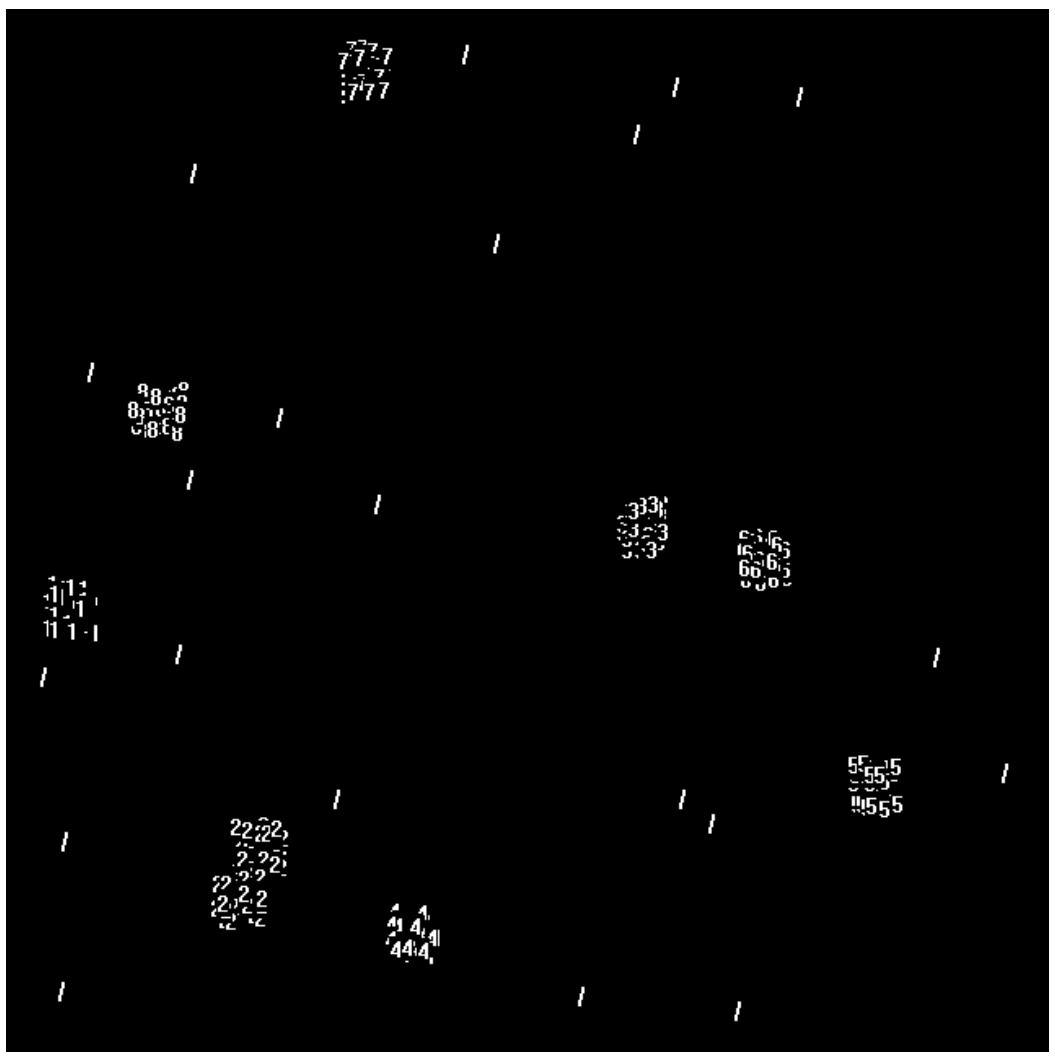

(f) Clusters identified by DBSCAN (8 clusers)

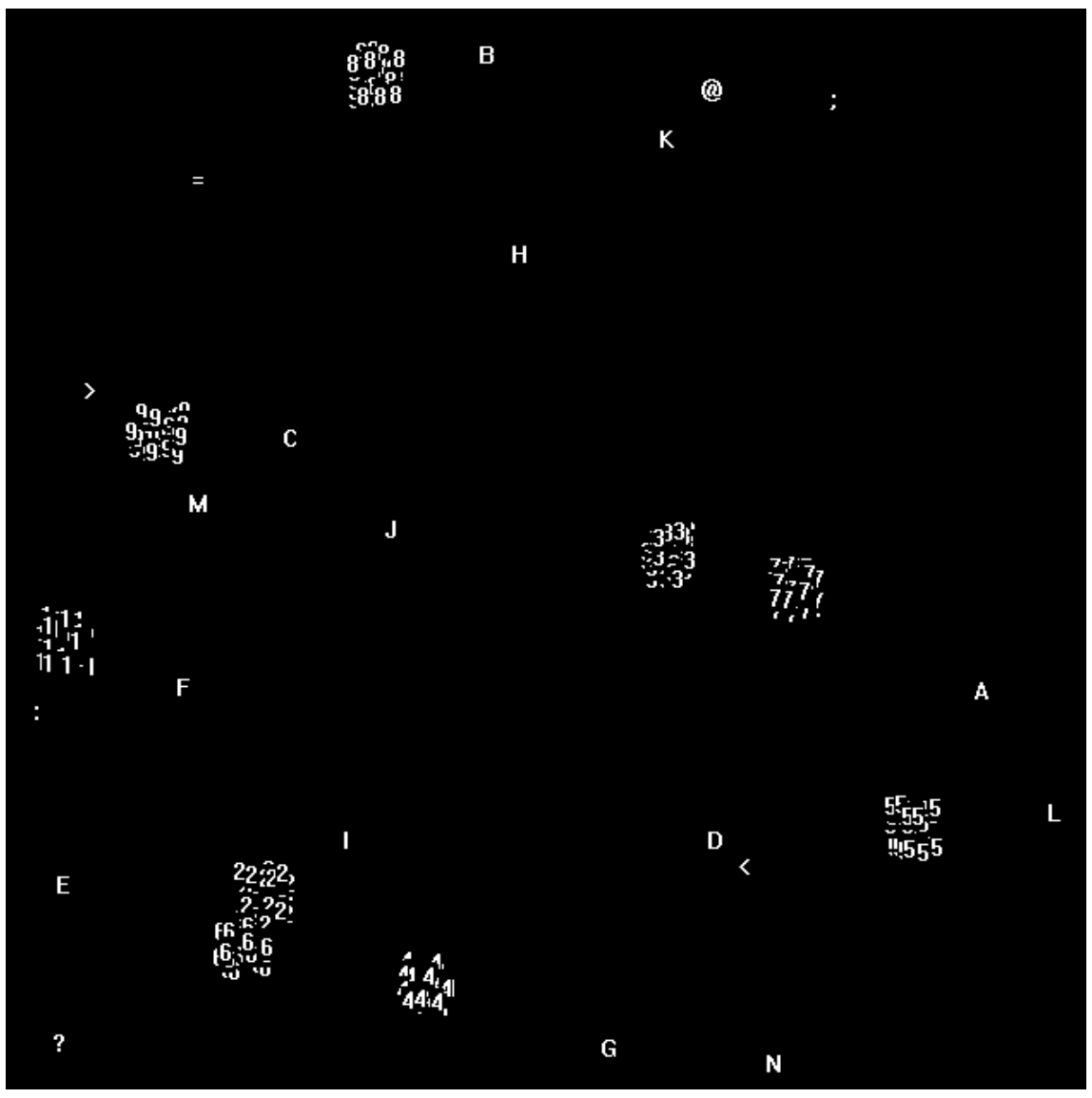

(g) Clusters identified by MeanShiftCluster (30 clusers)

Figure 7. Comparison of clustering results of several algorithms (DS3, $n = 400$)

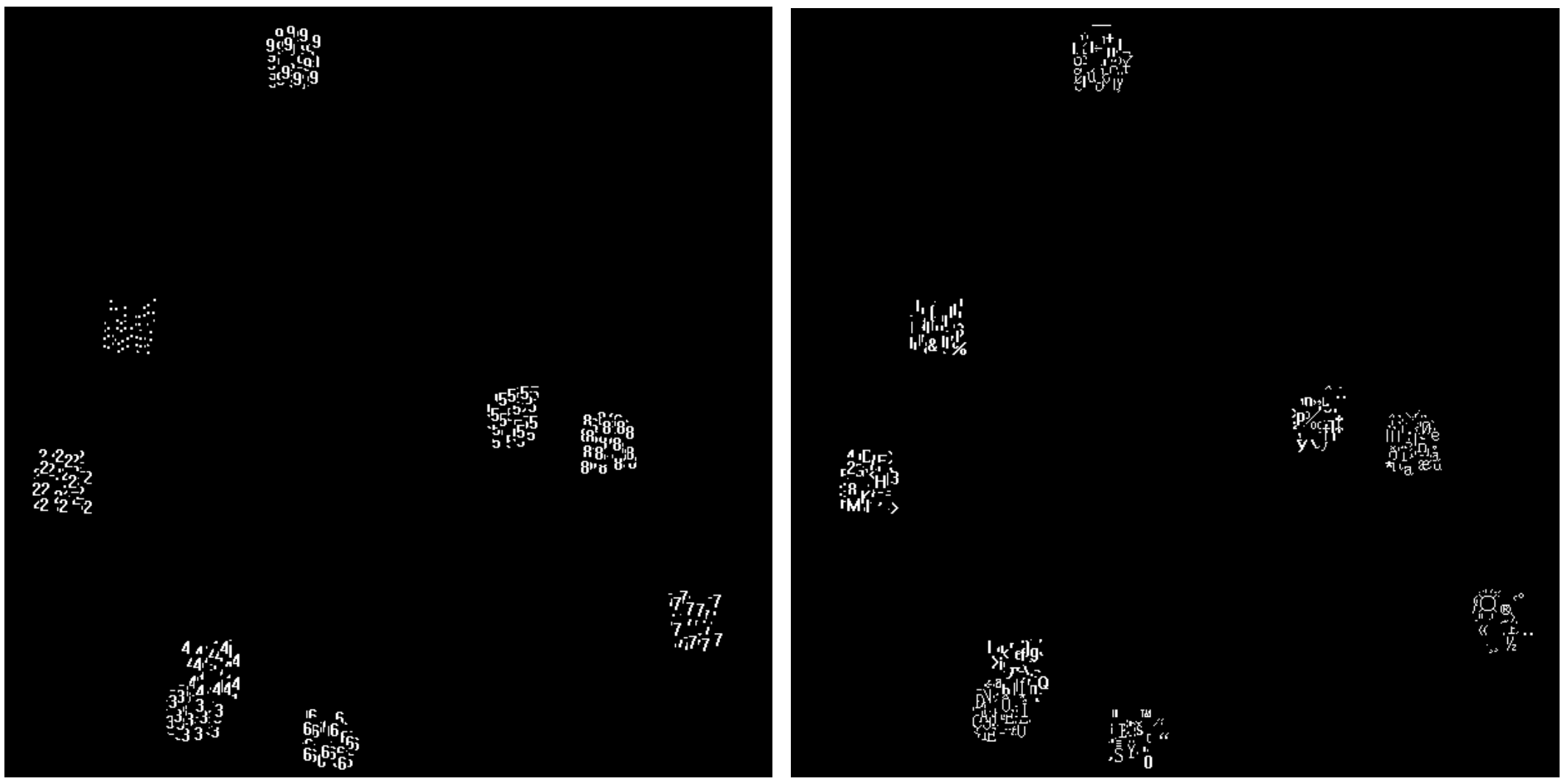

(a) Clusters identified by ESynC, IESynC, and SSynC (9 clusers) (b) Clusters identified by SynC (255 clusers)

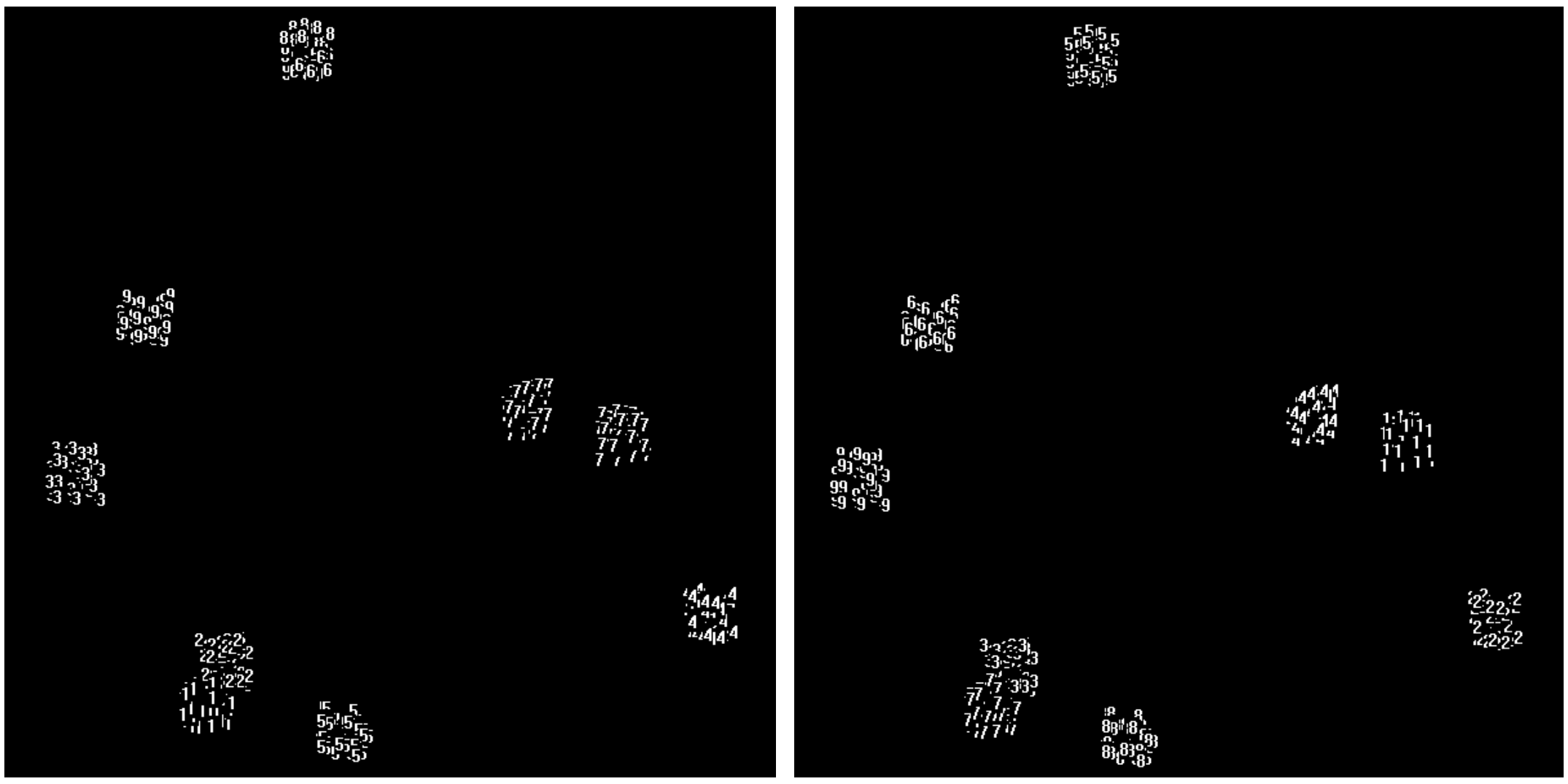

(c) Clusters identified by Kmeans (predefined 9 clusers) (d) Clusters identified by FCM (predefined 9 clusers)

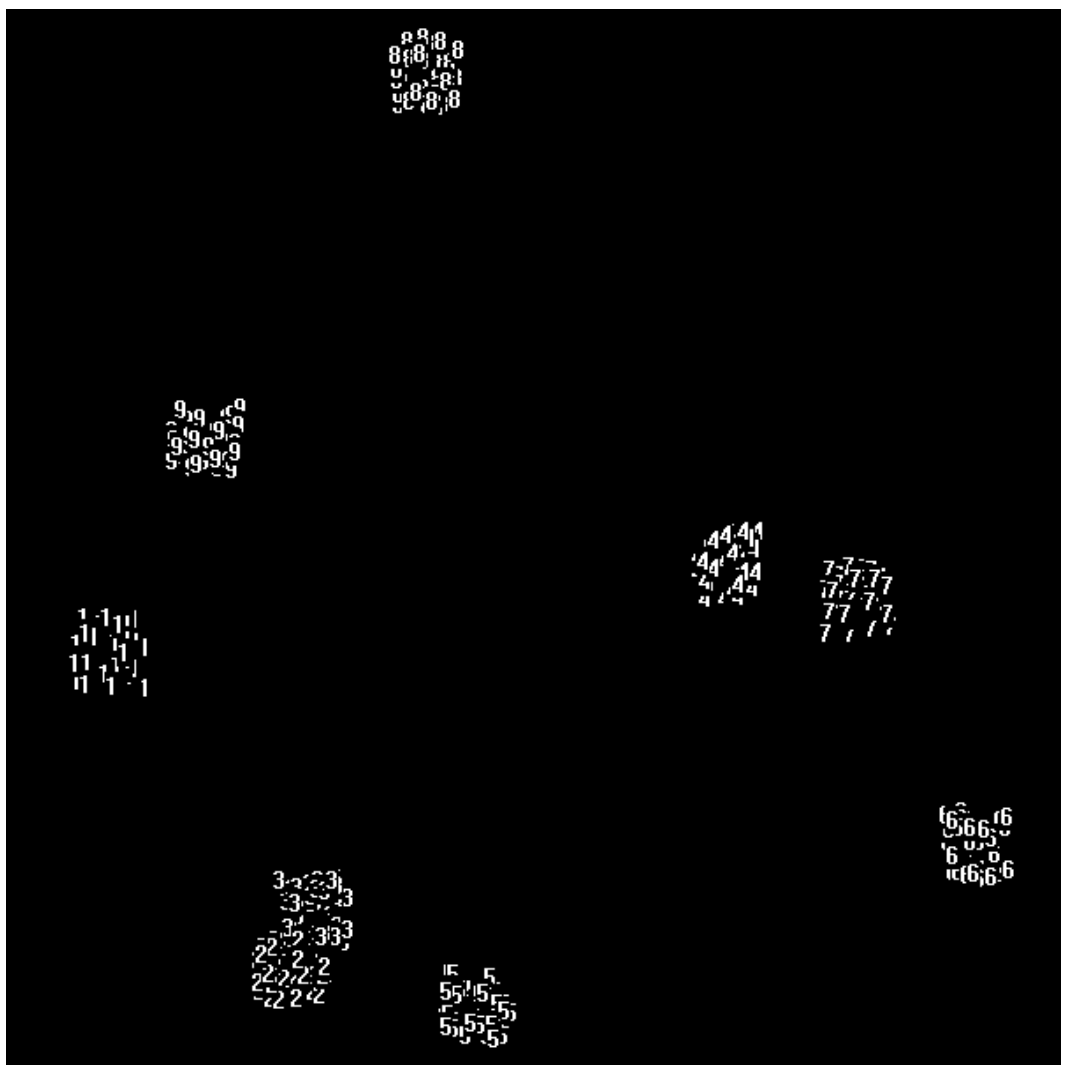

(e) Clusters identified by AP (9 clusers)

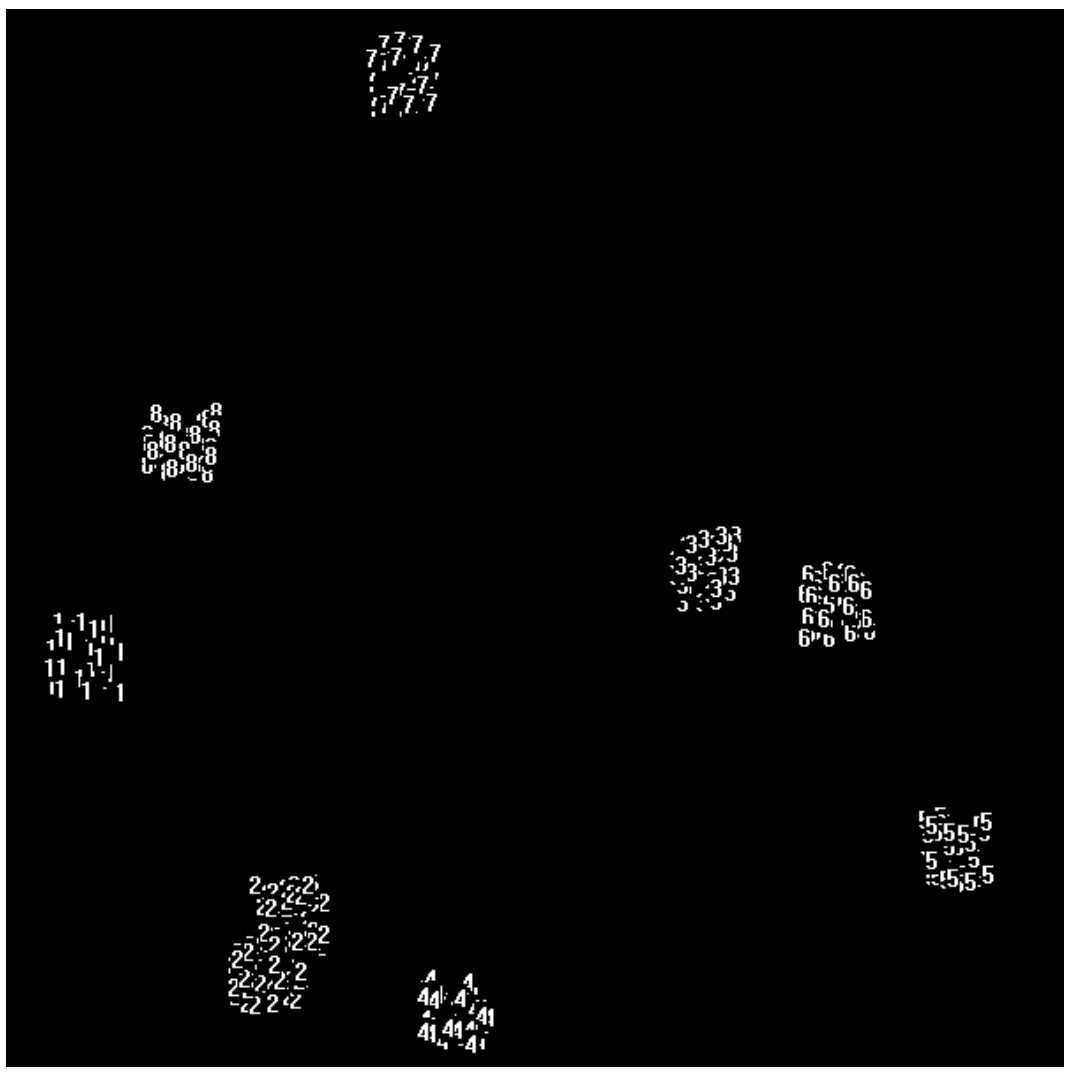

(f) Clusters identified by DBSCAN (8 clusers)

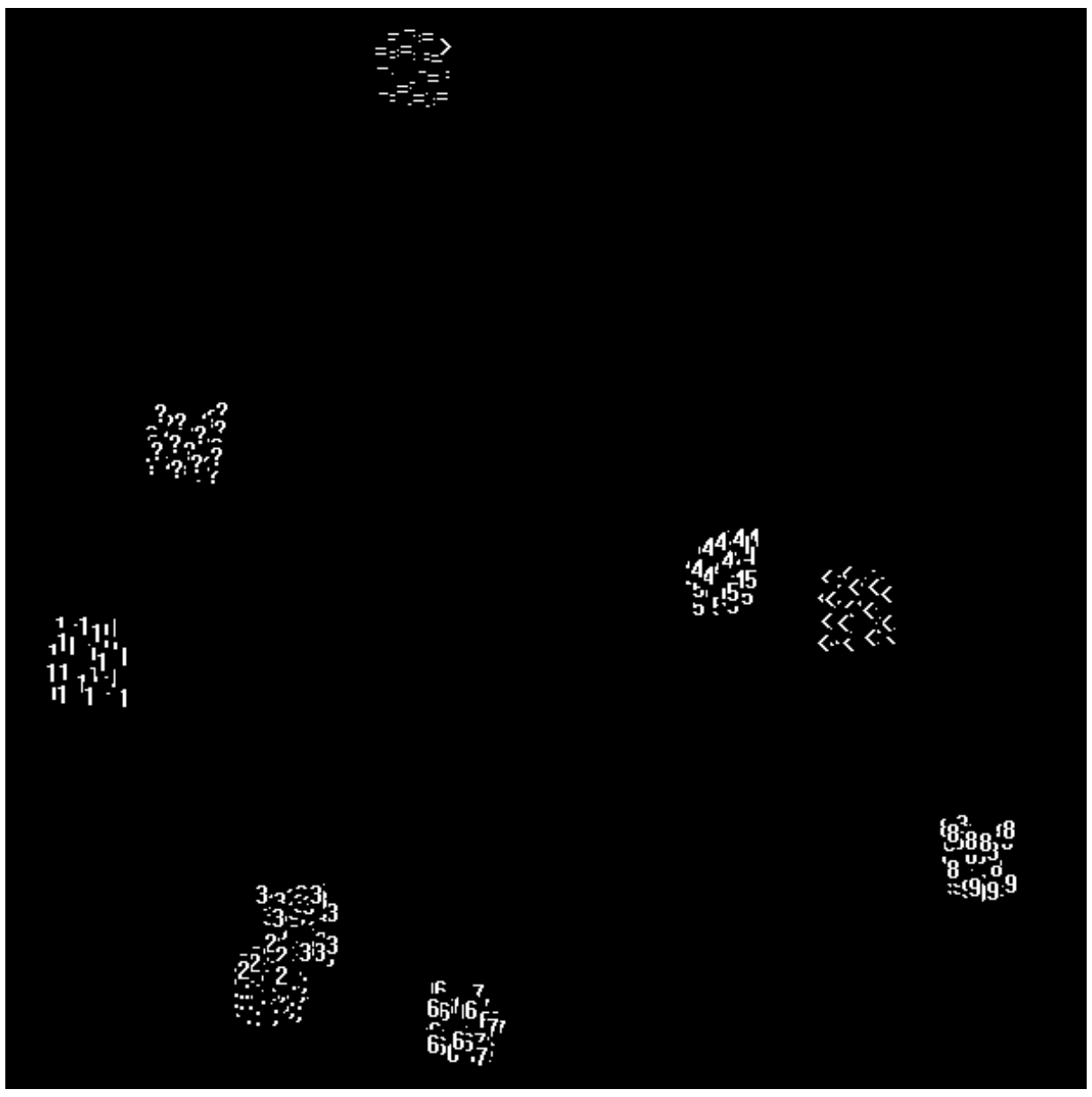

(g) Clusters identified by MeanShiftCluster (15 clusers)

Figure 8. Comparison of clustering results of several algorithms (DS4, $n$ = 400)

In simulations of some data sets, we find that the iterate times of SynC, AP, Kmeans, and FCM are larger than the iterate times of ESynC, IESynC, and SSynC. ESynC, IESynC, SSynC, and DBSCAN algorithms have better ability in exploring clusters and isolates than SynC, Kmeans, FCM, AP, and MeanShiftCluster algorithms for many kinds of data sets. Specially, AP algorithm needs the largest time cost.

From Fig. 5 to Fig. 8, parameter $\delta = 18$, the number of data points $n = 400$, parameter $\varepsilon$ (a very small real, if the distance of two points is less than $\varepsilon$, then they are regarded as in the same cluster; it also used as the iterative exit threash) = 0.00001. In IESynC, $r_1, r_2 = 20$.

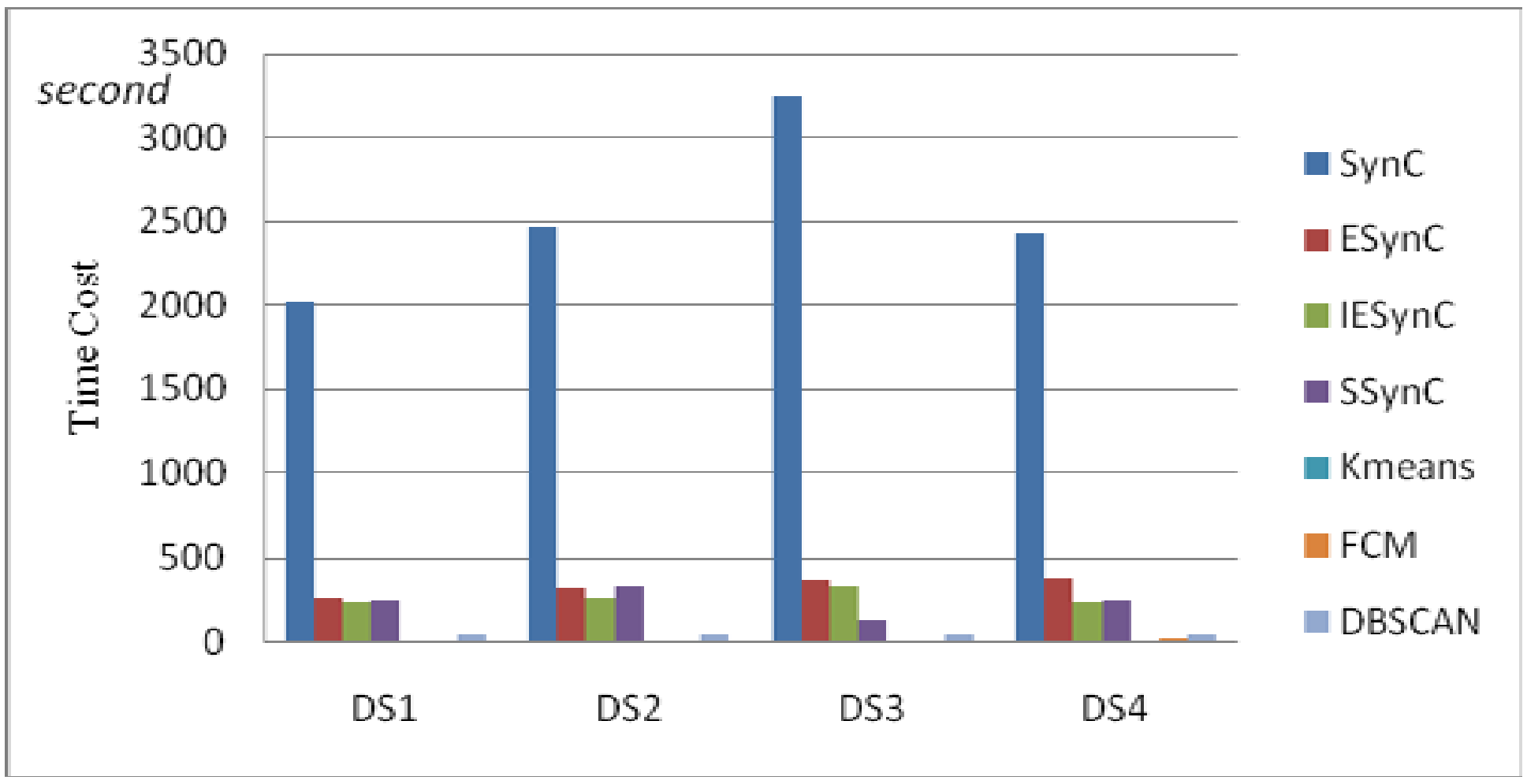

Figure 9. Comparison of several algorithms in time cost by using four artificial data sets (DS1 – DS4, $n$ = 20000).

In Fig. 9, parameter $\delta$ = 18, the number of data points $n$ = 20000, parameter $\varepsilon$ (a very small real, if the distance of two points is less than $\varepsilon$, then they are regarded as in the same cluster; it also used as the iterative exit threash) = 0.00001. In IESynC, $r_1, r_2$ = 6.

Figure 5 - Figure 8 are four comparative figures of clustering quality of several clustering algorithms using four data sets (from DS1 – DS4). From these figures, we find that ESynC, IESynC, and SSynC can get better clustering quality (obvious clusters or isolates displayed by figures) than SynC, AP, KMeans and FCM for some artificial data sets (from DS1 – DS4). Mean-Shift, DBSCAN can obtain similar clustering quality (obvious clusters displayed by figures) with ESynC, IESynC, and SSynC for some artificial data sets (from DS1 – DS4).

Especially, SynC, ESynC, IESynC, and SSynC can all easily find some isolatets if setting proper parameter $\delta$.

In some displayed figures, by intercomparing SynC, ESynC, IESynC, and SSynC, we find that SSynC can explore the same clusters and isolates (displayed by figures) with EsynC and IESynC. For many kinds of data sets, ESynC, IESynC, and SSynC can explore obvious clusters or isolates if setting proper parameter $\delta$, and SynC can not explore obvious clusters very well.

In Figure 9, Intercomparing ESynC, IESynC, SSynC, Mean-Shift, DBSCAN, FCM, and KMeans, we find that IESynC and SSynC is faster than ESynC with the same clustering results (the same clusters displayed by figures), and Kmeans and FCM are faster than other algorithms .

## 7.3 Experimental Results of Some Artificial Data Sets (DS5 – DS12)

Here we only compareSynC algorithm, ESynC algorithm, IESynC algorithm, SSynC algorithm, and DBSCAN algorithm in time cost and the number of clusters. From simulations, we observe that DBSCAN algorithm is fast obviously when the number of points is not very large. When parameter $\delta$ is set large, then SSynC algorithm can get similar time cost with DBSCAN algorithm.

Table 3 is the comparison results among SynC algorithm, ESynC algorithm, IESynC algorithm, and SSynC algorithm. The experiments of DS9 – DS12 can not operated efficiently in our computer because the number of grid cells is too large.

Table 3. Comparison of time cost for several different values of parameter $\delta$ among SynC, ESynC, IESynC, and SSynC by using several artificial data sets with variant dimensions. In Table 3, $n$ = 10000, parameter $\varepsilon$ (a very small real, if the distance of two points is less than $\varepsilon$, then they are regarded as in the same cluster; it also used as the iterative exit threash) = 0.00001.

(a). DS5 (In FSync, $N$ = 1404)

In IESynC algorithm, $r_i$ ($i$ = 1, 2) = 16. The effective region of parameter $\delta$ in ESynC, IESynC, and SSynC is [9, 58], and the effective region of parameter $\delta$ in DBSCAN is [2, 45].

| $\delta$ | **Spend time (*second*)** | | | | | **The number of clusters** | | |
|---|---|---|---|---|---|---|---|---|
| | SynC | ESynC | IESynC | SSynC | DBSCAN | SynC | ESynC<br>IESynC<br>SSynC | DBSCAN |
| 1 | | | | | 12 | | | 627 |
| 2 | | | | | 13 | | | 11 |
| 3 | | | | | 12 | | | 11 |
| 4 | | | | | 11 | | | 11 |
| 6 | | | | | 12 | | | 11 |
| 8 | | | 261 | | 11 | | 15 | 11 |
| 9 | | | 77 | | | | 12 | |
| 10 | | | 63 | | 11 | | 13 | 11 |
| 12 | 610 | 128 | 61 | 68 | 12 | 342 | 13 | 11 |
| 14 | 601 | 96 | 46 | 62 | 12 | 349 | 12 | 11 |
| 16 | 612 | 802 | 398 | 49 | 12 | 335 | 13 | 11 |
| 18 | 613 | 192 | 97 | 41 | 12 | 343 | 11 | 11 |
| 20 | 619 | 98 | 48 | 40 | 12 | 345 | 11 | 11 |
| 22 | 624 | 81 | 41 | 39 | 12 | 334 | 11 | 11 |
| 24 | 626 | 80 | 41 | 36 | 12 | 349 | 11 | 11 |
| 26 | 627 | 65 | 33 | 33 | 12 | 344 | 11 | 11 |
| 28 | 629 | 65 | 32 | 29 | 12 | 345 | 11 | 11 |
| 30 | 630 | 65 | 34 | 25 | 12 | 345 | 11 | 11 |
| 40 | | 49 | 25 | 17 | 12 | | 11 | 11 |
| 42 | | | | | 12 | | | 11 |
| 44 | | | | | 12 | | | 11 |
| 45 | | | | | 12 | | | 11 |
| 46 | | | | | 14 | | | 10 |
| 50 | | 48 | 25 | 16 | 11 | | 11 | 10 |
| 55 | | 64 | 33 | 16 | 11 | | 11 | 10 |
| 58 | | | | 16 | | | 12 | |
| 59 | | | | 16 | | | 10 | |
| 60 | | | | 16 | 12 | | 10 | 10 |

(b). DS6 (In FSync, $N$ = 16848)

In IESynC algorithm, $r_i$ ($i$ = 1, 2, 3, 4) = 50. The effective region of parameter $\delta$ in ESynC, IESynC, and SSynC is [11, 164], and the effective region of parameter $\delta$ in DBSCAN is [7, 147].

| $\delta$ | **Spend time (*second*)** | | | | | **The number of clusters** | | |
|---|---|---|---|---|---|---|---|---|
| | SynC | ESynC | IESynC | SSynC | DBSCAN | SynC | ESynC<br>IESynC<br>SSynC | DBSCAN |
| 6 | | | | | 17 | | | 45 |
| 7 | | | | | 17 | | | 12 |
| 8 | | | | | 18 | | | 12 |
| 10 | | | | 158 | 18 | | 15 | 12 |
| 11 | | | | 166 | | | 12 | |
| 12 | 860 | 135 | 92 | 147 | 18 | 5722 | 12 | 12 |
| 16 | 877 | 110 | 84 | 83 | 19 | 5694 | 12 | 12 |
| 20 | 883 | 89 | 76 | 69 | 17 | 5636 | 12 | 12 |
| 24 | 772 | 67 | 68 | 53 | 18 | 5629 | 12 | 12 |
| 28 | 784 | 67 | 69 | 52 | 18 | 5649 | 12 | 12 |
| 32 | 416 | 66 | 70 | 51 | 18 | 5584 | 12 | 12 |
| 36 | 419 | 67 | 69 | 42 | 17 | 5578 | 12 | 12 |
| 40 | 453 | 66 | 68 | 30 | 17 | 5566 | 12 | 12 |
| 50 | 421 | 67 | 68 | 21 | 17 | 5549 | 12 | 12 |
| 60 | 436 | 44 | 63 | 21 | 17 | 5548 | 12 | 12 |
| 80 | 436 | 45 | 63 | 21 | 18 | 5548 | 12 | 12 |
| 100 | 436 | 45 | 63 | 21 | 17 | 5548 | 12 | 12 |
| 140 | 436 | 44 | 74 | 21 | 17 | 5548 | 12 | 12 |
| 145 | | | | | 20 | | | 12 |
| 146 | | | | | 19 | | | 12 |
| 147 | | | | | 20 | | | 12 |
| 148 | | | | | 20 | | | 11 |
| 150 | | | | | 19 | | | 10 |
| 160 | 804 | 89 | 124 | 22 | 18 | 5550 | 12 | 9 |
| 162 | | | | 22 | | | 13 | |
| 164 | | | 1147 | 21 | 18 | | 12 | 8 |
| 165 | | | | 22 | | | 10 | |
| 170 | | 360 | 374 | 23 | | | 9 | |
| 180 | | 112 | | | 17 | | 8 | 8 |

(c). DS7 (In FSync, $N = 175616$)

In IESynC algorithm, $r_i$ ($i = 1, 2, 3, 4, 5, 6$) = 80. The effective region of parameter $\delta$ in ESynC, IESynC, and SSynC is [16, 214], and the effective region of parameter $\delta$ in DBSCAN is [12, 199].

| $\delta$ | **Spend time (*second*)** | | | | **The number of clusters** | | **The iterate times** | |
|---|---|---|---|---|---|---|---|---|
| | ESynC | IESynC | SSynC | DBSCAN | ESynC IESynC SSynC | DBSCAN | ESynC IESynC | SSynC |
| 10 | | | | 20 | | 133 | | |
| 11 | | | | 21 | | 29 | | |
| 12 | | | | 20 | | 12 | | |
| 14 | | | 191 | | 18 | | | 7 |
| 15 | | | 152 | | 14 | | | 6 |
| 16 | 135 | 50 | 151 | 20 | 12 | 12 | 5 | 5 |
| 20 | 109 | 42 | 111 | 21 | 12 | 12 | 4 | 4 |
| 100 | 45 | - | 24 | 20 | 12 | 12 | 2 | 2 |
| 180 | 45 | | 23 | 20 | 12 | 12 | 2 | 2 |
| 190 | | | | 20 | | 12 | | |
| 196 | | | | 20 | | 12 | | |
| 198 | | | | 20 | | 12 | | |
| 199 | | | | 20 | | 12 | | |
| 200 | 68 | | 24 | 20 | 12 | 11 | 3 | 3 |
| 210 | | | 23 | | 12 | | | 3 |
| 212 | | | 24 | | 12 | | | 3 |
| 213 | 90 | | 24 | | 12 | | 4 | 4 |
| 214 | 91 | | 24 | | 12 | | 4 | 4 |
| 215 | 1130 | | 28 | | 13 | | 50 | 50 |
| 216 | | | 23 | | 13 | | | 50 |
| 220 | | | 23 | 20 | 11 | 11 | | 7 |
| 260 | | | | 20 | | 10 | | |

(d). DS8 (In FSync, $N = 27648$)

In IESynC algorithm, $r_i$ ($i$ = 1, 2, 3, 4, 5, 6, 7, 8) = 180. The effective region of parameter $\delta$ in ESynC, IESynC, and SSynC is [22, 298], and the effective region of parameter $\delta$ in DBSCAN is [17, 281].

| $\delta$ | **Spend time (*second* )** | | | | **The number of clusters** | | **The iterate times** | |
|---|---|---|---|---|---|---|---|---|
| | ESynC | IESynC | SSynC | DBSCAN | ESynC IESynC SSynC | DBSCAN | ESynC IESynC | SSynC |
| 14 | | | | 24 | | 117 | | |
| 16 | | | | 24 | | 13 | | |
| 17 | | | | 23 | | 12 | | |
| 18 | | 109 | | 24 | 44 | 12 | 6 | |
| 20 | | 94 | | | 17 | | 5 | |
| 21 | | 94 | | | 14 | | 5 | |
| 22 | 118 | 79 | 121 | 23 | 12 | 12 | 4 | 4 |
| 24 | 119 | 81 | 121 | 23 | 12 | 12 | 4 | 4 |
| 100 | 60 | 50 | 30 | 24 | 12 | 12 | 2 | 2 |
| 180 | | | | | | | | |
| 250 | | | 30 | 24 | 12 | 12 | | 2 |
| 280 | | | 31 | 23 | 12 | 12 | | 2 |
| 281 | | | | 23 | | 12 | | |
| 282 | | | | 24 | | 11 | | |
| 286 | | | | 24 | | 11 | | |
| 290 | | | 31 | 24 | 12 | 11 | | 3 |
| 296 | | | 30 | | 12 | | | 4 |
| 298 | 119 | - | 31 | | 12 | | 4 | 4 |
| 299 | | | 30 | | 11 | | | 49 |
| 300 | | | 31 | 23 | 11 | 11 | | 33 |

## 6.5 Analysis and Conclusions of Experimental Results

From the comparative experimental results of Figure 5 - Figure 9, and Table 2 - Table 3, we observe that IESynC algorithm and SSynC algorithm are faster than ESynC algorithm for many cases. Usually, IESynC algorithm spend less time for many kinds of data sets by selecting a proper parameter $r_i$ ($i$ = 1, 2, …, $d$). In simulations of four data sets (from DS5 – DS8), we observe that the effective region of parameter $\delta$ in ESynC, IESynC, and SSynC is longer than the effective region of parameter $\delta$ in DBSCAN.

IESynC algorithm is an improved clustering algorithm with faster clustering speed than ESynC algorithm for many cases. The time cost of IESynC algorithm is sensitive to parameter $r_i$ ($i$ = 1, 2, …, $d$). Usually, if the data sets have obvious clusters, the number of grid cells is better approach the number of points. If the number of grid cells is too less or too large, perhaps IESynC algorithm can not obtain obvious improvements in time cost.

SSynC algorithm is another improved clustering algorithm with faster clustering speed than ESynC algorithm for many cases.

## 8. Conclusions

This paper presents another synchronization clustering algorithm, ESynC, which ofter gets better clustering results than the original synchronization clustering algorithm, SynC. From the view of the theoretic analysis, we can see that ESynC algorithm has less iterative times than SynC algorithm. From the experimental results, we observed that ESynC algorithm can often obtain faster clustering speed and better clustering quality than SynC algorithm for many kinds of data sets.

ESynC algorithm is also robust to outliers and can find obvious clusters with different shapes like SynC algorithm. The number of clusters does not have to be fixed before clustering. Usually, parameter $\delta$ has some valid interval that can be determined by using an exploring method listed in [19] or using the same method presented in [1]. In the process of constructing $\delta$ near neighbor point sets, the time cost of FSynC algorithm can often be decreased for many kinds of data sets by combining grid cell partitioning method and Red-Black tree index structure.

The ESynC algorithm has some similarity with Mean shift algorithm [25, 26], but they have basic difference. ESynC algorithm is a dynamic synchronization clustering algorithm. And Mean shift algorithm a non-parametric mode clustering algorithm that can be described by:

"The main idea behind mean shift is to treat the points in the *d*-dimensional feature space as an empirical probability density function where dense regions in the feature space correspond to the local maxima or modes of the underlying distribution. For each data point in the feature space, one performs a gradient ascent procedure on the local estimated density until convergence. The stationary points of this procedure represent the modes of the distribution. Furthermore, the data points associated (at least approximately) with the same stationary point are considered members of the same cluster." [26]

The next work is to explore the relation between ESynC algorithm and Mean shift algorithm further [25, 26] and to compare ESynC algorithm with SyncStream algorithm [27] in data stream mining.